\documentclass{article}

 \PassOptionsToPackage{numbers,square,sort&compress}{natbib}
\usepackage[preprint]{neurips_2026}

\usepackage[utf8]{inputenc} % allow utf-8 input
\usepackage[T1]{fontenc}    % use 8-bit T1 fonts
\usepackage{hyperref}       % hyperlinks
\usepackage{url}            % simple URL typesetting
\usepackage{booktabs}       % professional-quality tables
\usepackage{amsfonts}       % blackboard math symbols
\usepackage{nicefrac}       % compact symbols for 1/2, etc.
\usepackage{microtype}      % microtypography
\usepackage{xcolor}         % colors

\usepackage{tabularx}
\usepackage{array}

\usepackage[toc,page,header]{appendix}
\usepackage{minitoc}

 \usepackage{subcaption}

 \usepackage{caption}
\usepackage{hyperref}
\usepackage{url}
\usepackage{amsmath} 
\usepackage{graphicx} 
\usepackage{wrapfig}

\def\({\left(}
\def\){\right)}
\def\[{\left[}
\def\]{\right]}
\def\<{\langle}
\def\>{\rangle}
\newcommand{\bmat}{\begin{bmatrix}}
\newcommand{\emat}{\end{bmatrix}}

\def\tr{\mathop{\rm tr}}

\newcommand\half{{\ensuremath{\frac{1}{2}}}}
\newcommand\p{\ensuremath{\partial}}

\newcommand{\be}{\begin{equation}}
\newcommand{\ee}{\end{equation}}
\newcommand{\bea}{\begin{eqnarray}}
\newcommand{\eea}{\end{eqnarray}}
\newcommand{\bwt}{\begin{widetext}}
\newcommand{\ewt}{\end{widetext}}

\newcommand{\bi}{\begin{itemize}}
\newcommand{\ei}{\end{itemize}}
\newcommand{\ben}{\begin{enumerate}}
\newcommand{\een}{\end{enumerate}}
\newcommand{\bca}{\begin{cases}}
\newcommand{\eca}{\end{cases}}
\newcommand{\bln}{\begin{align}}
\newcommand{\eln}{\end{align}}
\newcommand{\bst}{\begin{split}}
\newcommand{\est}{\end{split}}

\newcommand\al{{\alpha}}
\newcommand\ep{\epsilon}
\newcommand\sig{\sigma}
\newcommand\Sig{\Sigma}
\newcommand\lam{\lambda}

\newcommand\om{\omega}

\def\th{{\theta}}

\newcommand\ha{{\half}}

\def\le{\left}
\def\ri{\right}

\newcommand\sO{{\ensuremath{{\mathcal O}}}}

\newcommand{\NI}[1]{\textcolor{blue}{\textsf{[NI: #1]}}}

\title{Spontaneous symmetry breaking and Goldstone modes for deep information propagation}

\author{%
  \hspace{-0.5cm}Nabil Iqbal\thanks{Equal contribution.} 
  \\
  Dept. of Mathematical Sciences, Durham University\\
  AMLab, University of Amsterdam\\
  \And
   T. Anderson Keller\footnotemark[1] \\
  Kempner Institute, Harvard University \\
  \AND
  \hspace{1cm} Yue Song \\
  \hspace{0.05\linewidth}College of AI, Tsinghua University \\
  \And
 \hspace{1.1cm}Takeru Miyato \\
  \hspace{1.1cm}University of Tübingen, Tübingen AI Center \\
  \AND
  Max Welling \\
  CuspAI \\
  AMLab, University of Amsterdam \\
}

\begin{document}
\doparttoc % Tell to minitoc to generate a toc for the parts
\faketableofcontents % Run a fake tableofcontents command for the partocs
\part{} % Start the document part
\vspace{-10mm}
% \parttoc % Insert the document TOC

\maketitle

\begin{abstract}
\looseness=-1
In physical systems, whenever a continuous symmetry is spontaneously broken, the system possesses excitations called Goldstone modes, which allow coherent information propagation over long distances and times. In this work, we study deep neural networks whose internal layers are equivariant under a continuous symmetry and may therefore support analogous Goldstone-like degrees of freedom. We demonstrate, both analytically and empirically, that these degrees of freedom enable coherent signal propagation across depth and recurrent iterations, providing a mechanism for stable information flow without relying on architectural stabilizers such as residual connections or normalization. In feedforward networks, this results in improved trainability and representational diversity across layers. In recurrent settings, we demonstrate the same mechanism is valuable for long-term memory by propagating information over recurrent iterations, thereby improving performance of RNNs and GRUs on long-sequence modeling tasks.

\end{abstract}

\section{Motivation}
The flow of information through the layers of a neural network is central to deep learning and has been the topic of intense study.  It is interesting to consider the layer direction to be analogous to ``time'' in a physical system. In that case, the problem of information flow through the network is mapped to the problem of creating stable structures in nature in a potentially turbulent environment. One can ask how this problem is solved in physical systems. This analogy becomes even sharper if one considers networks with a convolutional or recurrent architecture, which enforce the physical constraints of locality and translational invariance in space and time respectively. 

\looseness=-1
In fact, there is a universal physical principle which allows information to propagate coherently across long distances in space and time, that of {\it spontaneously broken symmetry}. Whenever the dynamics of a system are invariant under a symmetry operation, but the equilibrium configuration of the system is {\it not} invariant under the same symmetry, as shown in Figure \ref{fig:potential}, we say that the symmetry is {\it spontaneously broken}. As we briefly review in Appendix \ref{app:goldstone}, it is well-known that in situations of broken continuous symmetry the system generally possesses long-lived excitations called {\it Goldstone modes}, corresponding to coherent modulations of the equilibrium state by the symmetry operation \cite{Goldstone:1962es}. 

Such Goldstone modes play a central role in physics, and there is a sense in which much of the long-distance information propagation in nature (e.g. sound \cite{Leutwyler:1996er}, spin waves in magnets \cite{sachdev1999quantum}, the pion exchange which mediates nuclear structure \cite{Weinberg:1978kz}, even a modern understanding of light \cite{Gaiotto:2014kfa,Lake:2018dqm,Hofman:2018lfz}) can be related to a spontaneously broken symmetry. In this work we consider a deep-learning analogue of this phenomenon. In particular, we demonstrate that in neural networks where {\it internal layers} are equivariant under a symmetry, degrees of freedom which are somewhat analogous to Goldstone modes propagate through the network, carrying information. These modes are outside the conventional ``edge of chaos'' paradigm for deep neural networks, which assume no extra structure. We show that they lead to concrete advantages in neural network performance, allowing very deep networks to be trained with no extra mitigations (e.g. LayerNorm, skip connections, etc.) and providing benefits in the performance of recurrent neural networks on long-sequence tasks. 

\subsection{Goldstone modes in physics} 
We provide some very brief motivational remarks about symmetries and their spontaneous breaking in the form of an example. Consider a system described by a complex variable $\phi$, whose potential energy is given by a function $V(|\phi|) = r|\phi|^2 + \lam |\phi|^4$. Note this potential energy is invariant under the phase rotation $\phi \to e^{i\al} \phi$. The equilibrium configurations of this system are found by minimizing the potential energy. The structure of these low-energy configurations qualitatively differs depending on the sign of $r$. If $r>0$, there is only one unique solution $\phi = 0$ which minimizes the energy. This configuration is invariant under the symmetry operation, and we say that the symmetry is {\bf unbroken}. However if $r < 0$ then we find a {\it ring} of potential solutions with $|\phi| = \sqrt{\tfrac{-r}{2\lam}}$. The solution must pick one point along this ring, and it is no longer invariant under the symmetry operation, which now takes us from one solution to another. In this situation we say that the symmetry is {\bf spontaneously broken}. 
\begin{figure}
    \centering
    \vspace{-10mm}
    \includegraphics[width=0.7\linewidth]{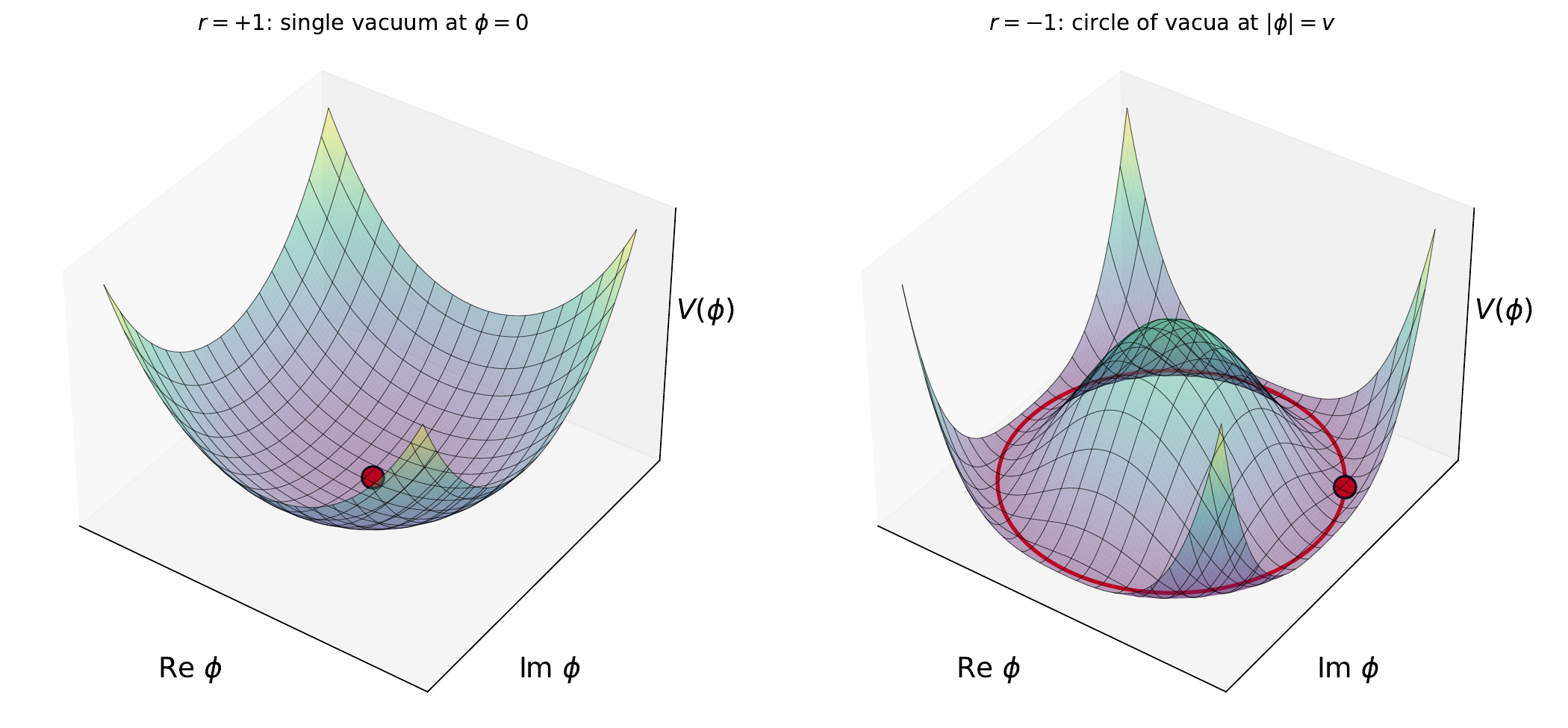}
    \caption{Equilibrium configurations are minima of the potential energy $V(\phi)$. If $r > 0$ we have a single isolated minimum at $\phi = 0$ and the symmetry is {\bf unbroken}. If $r < 0$ we find a ring of minima with constant $|\phi|$, of which the system must choose one; the symmetry is {\bf spontaneously broken}. }
    \vspace{-5mm}
    \label{fig:potential}
\end{figure}

\looseness=-1
The continuous family of low-energy solutions often results in dramatic changes to the long-time dynamics. In physics one often considers a situation where $\phi = \phi(t,x)$ is a {\it field} which depends on space and time, and then in the spontaneously broken phase one finds waves called {\bf Goldstone modes} which propagate coherently for long distances due to their intimate relation with the symmetry. We provide a brief review in Appendix \ref{app:goldstone}. 
In this work we find analogous phenomena in the context of deep learning.

\looseness=-1
\subsection{Prior work} 
Information propagation through the layers of a deep neural network, or over the timesteps of a recurrent network, is a very well-studied problem; see e.g. \cite{poole2016exponential,schoenholz2017deep,yang2017mean,xiao2018dynamical,chen2018dynamical}. We will extend this formalism to include equivariance, uncovering new dynamics. Our neural networks fit into a larger body of literature \cite{miyato2024artificial,liboni2023image, keller2023traveling} that is inspired by physical systems and recent observations in neuroscience \cite{miller_waves_memory, muller_sejnowski_waves_review, das_2026}. Group equivariance in neural networks \cite{cohen2016group} is an extremely mature field (see e.g. the textbooks \cite{ weiler2023equivariant,bronstein2021geometric}), though generally it is related to a symmetry of the {\it task}: here our tasks have no symmetry, and we consider equivariance as a purely architectural tool for information propagation. See \cite{kaba2023symmetry,smidt2021finding} for applications of symmetry breaking in the context of a given task in equivariant networks. In  \cite{lowe2022complex,lowe2024binding} equivariances similar to ours were used as an architectural tool, rather than as a symmetry of the data distribution. 

Related work in the context of a {\it discrete} symmetry rather than a continuous one includes \cite{iqbal2025topological}, where it was seen that topological defects play a role that is somewhat analogous to the Goldstone modes in our study. The advantages of a non-unique late-time attractor -- similar to the symmetry orbits that we find -- in the context of graph neural network oversmoothing were discussed in \cite{turan2026beyond}. See also the recent work \cite{mo2026symmetry}, which also discusses the influence of symmetry on dynamical stability. 

\section{Model and theoretical description}
\label{sec:model_theory}
We now describe our general framework for building deep neural networks which exhibit spontaneously broken symmetries, and provide preliminary analytical and empirical results that support the role of Goldstone modes in deep information propagation.
\subsection{Equivariant internal layers}
 We will consider deep neural networks with $L+2$ layers, of which the {\it bulk} of the network preserves a given symmetry group $G$. We denote the internal representation at layer $l$ by $\mathbf{x}^{l} \in \mathbb{R}^{N}$. We will sometimes need to discuss the individual components of $\mathbf{x}^l$, which we denote as $x^{l}_i$ with $i \in \{1, \cdots N\}$. 

Each internal state is acted on by a layer $f^{l}$ as
\be
\mathbf{x}^{l+1} = f^{l}(\mathbf{x}^{l}) \qquad f^{l}:\mathbb{R}^{N} \to \mathbb{R}^{N} \ . 
\ee
\looseness=-1
We assume that $\mathbf{x}^l$ transforms under some representation of $G$. We construct the internal layers $f^l$ to be equivariant under this transformation, i.e. for every $g \in G$ we have a representation matrix $\rho_{g}$ such that
\be
\rho_{g} f^{l}(\mathbf{x}^{l}) = f^{l}(\rho_{g} \mathbf{x}^{l}) \qquad l \in \{0,\cdots,L-1\}.
\label{equivlayers} 
\ee
This is straightforwardly done by appropriately restricting $f$. 

We now consider the input to the network $\mathbf{x}^\mathrm{in}$ and the output $\mathbf{x}^\mathrm{out}$. In most applications of a group-equivariant neural network \cite{cohen2016group}, $G$ corresponds to a symmetry of the {\it task}, and thus both the input and output have a well-defined transformation under $G$. In our framework, on the other hand, we use $G$-equivariance only as a tool for propagating and processing information. It has nothing to do with the tasks that we will study. Thus the first and final layers of the network $f^{\mathrm{in}}$ and $f^{\mathrm{out}}$ are fully general and break equivariance. We will see that the equivariance will still be useful for propagating information through layers.  

Some examples of $G$ that we will study include:
\begin{itemize}
\item $U(1)$: in this case it is convenient to consider the representation space to be $\mathbb{C}^{N}$, where the symmetry acts as 
\vspace{-3mm}
\be
z_{i} \to e^{i\al} z_{i} \label{U1trans} 
\ee
i.e. by a constant phase rotation on all the components of $\mathbf{z}$. 
\item $O(k)$: Here we consider a $k$ which divides $N$ and we consider $\mathbb{R}^{N}$ as a direct sum of $N / k$ blocks of $k$-dimensions each, each of which is acted on by the $O(k)$ matrix. Details are provided in Appendix \ref{app:equivariant-layers}. 
\end{itemize}
One can imagine that each unit acted on by $G$ (i.e. the complex entries in the $U(1)$ case, or the $k$-vectors in the $O(k)$ case), form a {\it capsule} \cite{sabour2017dynamic}, where in our case the capsule is defined in terms of its symmetry transformation.

%Concretely, in this section we will study a feedforward architecture, for which we have:
%\be
%x_{i}^{l+1} = \phi\left(\sum_{j} W_{ij}^l x^l_j\right)
%\ee
%where the weights and nonlinearity are chosen appropriately to preserve $G$-equivariance. We discuss the exact manner in which this is done for the various symmetries in \NI{appendix}. 

\subsection{Phases of symmetry and information propagation}
In this section we describe how symmetries affect information propagation. We use the $U(1)$ case for concreteness, but most of the results apply to any continuous symmetry group with the appropriate change of notation. 

In this section we disregard the non-equivariant input and output layers, and consider a $U(1)$-equivariant network whose updates take the form:
\be
\setlength{\abovedisplayskip}{4pt}
\setlength{\belowdisplayskip}{2pt}
z^{l+1}_{i} = \sum_{j} W^{l}_{ij} \phi(z^{l}_{j})  
\label{feedforwardU1} 
\ee
where we use a complex notation where at each layer $l$, $z_{i}^{l} \in \mathbb{C}$ and $W^{l}_{ij}$ is a complex matrix. The non-linearity is picked to be equivariant under a phase rotation, e.g. 
\be
\phi(z) = \frac{\tanh(|z|)}{|z|} z \label{nonl} 
\ee
which maps the magnitude of the complex number $|z|$ to $\tanh(|z|)$ while preserving its phase. This network is equivariant under a $U(1)$ symmetry acting on the inputs as \eqref{U1trans}\footnote{The theory that follows holds for any smooth and bounded activation function, though details such as the precise location of the phase transition line, the functional form of $c^{\star}$ etc. depend on the precise choice of activation.}

In this work, we interpret information propagation through three complementary perspectives: (i) preservation of input-dependent degrees of freedom, (ii) non-vanishing components of the input–output Jacobian, and (iii) sustained correlations across layers. We will relate these features to the possible phases of deep information propagation in the presence of symmetry. We do this by studying the behavior of the network at initialization using the tools developed in \cite{schoenholz2017deep,yang2017mean,xiao2018dynamical,chen2018dynamical}. 

We take each entry of $W_{ij}$ at initialization to be an i.i.d. complex number where the real and imaginary parts have variance $\frac{\sig_{W}^2}{N}$. We briefly recall the results of \cite{schoenholz2017deep} for a conventional feedforward network without any equivariance properties. The behavior of the network at initialization depends on the value of $\sig_{W}$: if this is small then the network will tend to collapse all inputs to a single attractor, thereby removing all information of the inputs. On the other hand, if it is very large then it will result in chaotic behavior which {\it effectively} decorrelates inputs from outputs. Both of these behaviors result in poor training, essentially not allowing information to propagate further than a certain finite (calculable) depth scale $\xi$. It is argued in \cite{schoenholz2017deep} that for efficient training one should instead tune $\sig_{W}$ to the precise boundary between these two behaviors, i.e. to the {\it edge of chaos}, at which $\xi$ diverges.

However the situation in the case of equivariance is fundamentally different. Consider performing a $U(1)$ rotation on an input $z^{0}_i \to e^{i\al} z^0_i$ to an $L$-layer equivariant network as in \eqref{U1trans}. Due to the equivariance the action of this rotation must be faithfully transmitted all the way through the layers of the network, rotating the output $z^{L}_i \to e^{i\al} z^{L}_i$. 

There are now two broad possibilities for what happens for very deep networks: 
\begin{enumerate}
\item {\bf Symmetry unbroken phase:} here we have $z^{L}_i \approx 0$, i.e. the network has forced all the inputs to zero. In this case the symmetry operation has no effect on the output. 
\item {\bf Symmetry spontaneously broken phase:} here we have $z^{L}_i \neq 0$, i.e. the final output has nonzero magnitude. In this case the symmetry necessarily acts non-trivially on the output.
\end{enumerate} 
The order parameter which distinguishes these two phases is: 
\be
\setlength{\abovedisplayskip}{3pt}
\setlength{\belowdisplayskip}{3pt}
c^{l} \equiv \frac{1}{N} \mathbb{E}\le(\sum_{j} \phi(z_j^l) \phi^{\dagger}(z_j^{l})\ri) \label{cstardef} 
\ee

\begin{wrapfigure}{l}{0.45\linewidth}
    \vspace{-0.6cm}
    \centering
    \includegraphics[width=\linewidth]{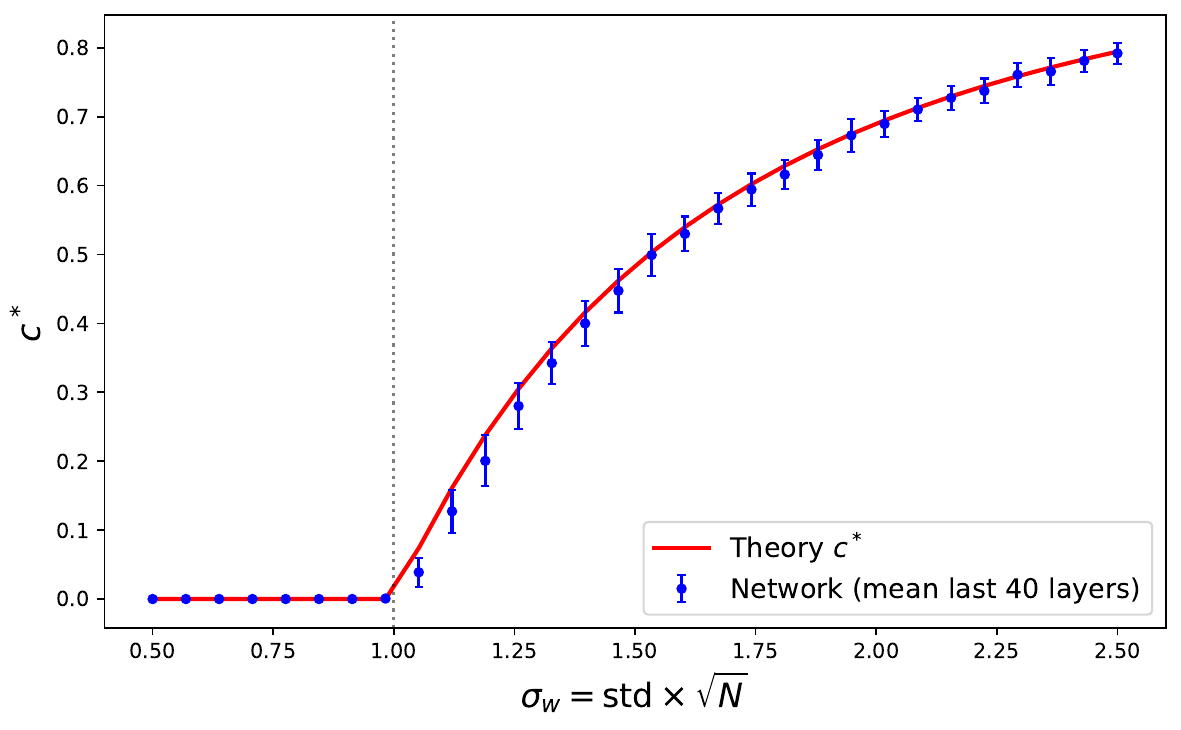}
    \caption{Comparison of mean-field result for $c^{\star}$ from numerically solving \eqref{cevolve} with empirical result measured over an ensemble of $L = 100, N = 16$ neural networks at initialization. Note clear phase transition at $\sig_{W} = 1$.}
    \label{fig:cstar}
    
\end{wrapfigure} 

i.e. the $U(1)$-invariant magnitude of each activation, averaged over the whole set, where the expectation value is measured over randomly sampled weights and some appropriate input distribution at initialization. If this $c^{l} \to 0$ as $l \to \infty$ we have the symmetry unbroken phase, and if it is finite then we have the Spontaneous Symmetry Broken (SSB) phase.

These two phases may appear to be in rough correspondence with the order and chaos phases in the usual discussion of deep information propagation \cite{schoenholz2017deep}. A difference is that the attractor in the ``order'' phase is necessarily the only $U(1)$-invariant point (i.e. $0$). Conversely, in the ``chaos'' phase the radius of each neuron (in expectation) approaches the calculable constant $\sqrt{c^{l}} > 0$, but the overall phase of the whole vector $\mathbf{z}$ is not set by the propagation over the layers, but rather by the $U(1)$ phase of the input. This is analogous to the choice of vacuum in the potential shown in the right panel of Figure \ref{fig:potential}. As a symmetry operation is faithfully transmitted through the network, the presence of equivariance then provides a stable path for some information to flow, stored in this overall phase. The presence of this channel may loosely be understood as a Goldstone mode.

%\YS{In our setting, the Goldstone mode corresponds to the global phase degree of freedom of the representation. This degree of freedom remains invariant across layers \eqref{phievolve} and induces a protected O(1) component in the input–output Jacobian \eqref{protected_jac}, thereby providing a coherent channel for information propagation.}

To formalize the idea of the flow of information, in Appendix \ref{app:jacobian} we show that a certain component of the Jacobian of the input-output map $\frac{\p z^{L}}{\p z^{0}}$ can be related to this order-parameter,
\be
d^{L} =  \frac{1}{N}\sum_{i}\mathbb{E}\le(\bigg|\sum_{j} \le(\frac{\p z^{L}_i}{\p z^{0}_j}  z^{0}_j - \frac{\p z^{L}_i}{\p z^{0\dagger}_j}  z^{0\dagger}_j\ri)\bigg|^2\ri) \label{protected_jac} 
\ee
where $d^{l}$ is the pre-activation analogue of $c^{l}$. This formula arises from considering how an infinitesimal symmetry transformation is propagated through the network. The left-hand side of this expression is $O(1)$ everywhere in the SSB phase: this means that a certain component of the Jacobian always remains $O(1)$, no matter how deep the network. This is non-trivial, as for vanilla networks these Jacobians tend to either collapse or explode. We will show below that this regularity seems to confer a concrete advantage in training very deep networks.  

%Note that in the second case, the information of how the symmetry acts must be faithfully transmitted through the layers. 

% \NI{we may be short on space: I can put this paragraph in an Appendix as its more of an apologia to physicists.} 
Finally, we note that though we have borrowed the terminology from equilibrium statistical mechanics in physics, the dynamics are slightly different. In the latter, there is no scope for considering the input/output map, which is our primary motivation. Conversely, in the latter, one usually has some notion of translational invariance in space that relates different degrees of freedom to each other, and order parameters generally measure how aligned these degrees of freedom are, with sharp transitions happening in the thermodynamic limit where we consider infinitely large systems. Thus spontaneous symmetry breaking is generally considered to be equivalent to the idea of {\it long-range order} (see e.g. \cite{chaikin1995principles}). In our current framework we do not generally have such an invariance: for an MLP processing a given data distribution, different neurons are not really related by a symmetry\footnote{This is interestingly different in e.g. the case of a convolutional or transformer architecture, where indeed different neurons {\it are} related by translational or permutation symmetry respectively.}, and we believe it is not really meaningful to ask whether distinct neurons are ``aligned''. Thus our order parameter is different from that in equilibrium statistical mechanics. %The intuition from Goldstone dynamics still carries over. 

\subsection{Large $N$ analysis}
\label{sec:large_N}
We now demonstrate that some aspects of the dynamics can be understood theoretically at initialization. Consider supplying an input $z^0_i$ at the start of the network. We study how the distribution of these inputs changes due to the influence of the randomly chosen $W_{ij}$ as we move through the network. In the large-$N$ limit it is well-known that this evolution is given by a Gaussian process \cite{schoenholz2017deep,lee2017deep}. We extend this calculation to our equivariant model in Appendix \ref{app:path-integral}, performing a self-contained derivation using the tools of stochastic path integrals. Here we provide only results.

\paragraph{Order parameter dynamics.}
In the large-$N$ limit the distribution of activations can be taken to be a Gaussian whose covariance $c^{l}$ defined in \eqref{cstardef} satisfies a well-defined evolution equation. The evolution of $c^{l}$ through layers is given by \eqref{app_cevolve} in the Appendix:
\be
c^{l+1} = 2 \int_{0}^{\infty} du~ u e^{-u^2} \le(\phi(u \sig_{W} \sqrt{c^{l}})^2\ri) \label{cevolve} 
\ee
From now on for concreteness we will specialize to the choice of non-linearity $\phi = \tanh$. Then there is a clear solution at $c^{l} = 0$, corresponding to all activations collapsed at zero. To see if this solution is stable, it is instructive to expand out the right-hand side of \eqref{cevolve} and perform the integral at small $c^{l}$ to find
\be
c^{l+1} = \sig_{W}^2 c^{l} + \sO((c^l)^2)
\ee
and thus we see that for $\sig_{W}^2 < 1$, $c^{l} \to 0$ as we take $l$ large. Here all activations are eventually zero and thus invariant under the symmetry: as described above this is the {\bf unbroken symmetry phase}. 

%\begin{figure}
%    \centering
%    \includegraphics[width=0.5\linewidth]{c_star_vs_sigma.pdf}
%    \caption{Comparison of mean-field result for $c^{\star}$ from numerically solving \eqref{cevolve} with empirical result measured over an ensemble of $L = 100, N = 16$ neural networks at initialization.}
%    \label{fig:cstar}
%\end{figure}

In contrast, for $\sig_{W}^2 > 1$, the solution at $c^{l} = 0$ is unstable and the system heads to a non-trivial fixed point. It is possible to numerically solve \eqref{cevolve} with $c^{l} = c^{l+1} = c^{\star}$ to find this fixed point, which we do in Figure \ref{fig:cstar}, comparing it with the value of $c^{\star}$ obtained from many instantiations of the network at initialization, finding excellent agreement. This is the {\bf symmetry spontaneously broken phase}. The phase transition is at $\sig_{W} = 1$.

\paragraph{Two-input correlation.} We now consider the covariance of two inputs $z^{l}_{a,b}$ where $a,b \in \{1, 2\}$. We now have two $U(1)$ symmetries: $U(1)_{a} \times U(1)_{b}$, and can construct the covariance matrix:
\be
C^l_{ab} \equiv \frac{1}{N} \mathbb{E}\le(\sum_{j} \phi(z_{j,a}^l) \phi^{\dagger}(z_{j,b}^{l})\ri) = \begin{pmatrix} 
            c^{l} & \Delta^{l} e^{i\varphi^{l}} \\
            \Delta^{l} e^{-i\varphi^{l}} & c 
            \end{pmatrix}
\ee
where the diagonal entries are the covariance $c^{l}$ that has previously been studied\footnote{It is possible in principle to have $C_{11} \neq C_{22}$; this is set by the initial conditions, i.e. the ensemble from which $z^0_{1}$ and $z^0_{2}$ are drawn. Here we assume for notational simplicity that $C_{11} = C_{22}$; the evolution will then not change this fact.} and the off-diagonal entries are now complex numbers which we write in terms of a magnitude $\Delta^{l}$ and phase $\varphi^{l}$. Using the same techniques as before we can write down a propagation equation for this covariance. There is now a crucial difference arising from the equivariance: we find that 
\be
\setlength{\abovedisplayskip}{3pt}
\setlength{\belowdisplayskip}{3pt}
\varphi^{l+1} = \varphi^{l} \label{phievolve} 
\ee
i.e. the {\it phase} of the 2-input covariance {\it does not change as we process through the layers}. This captures the idea that the $U(1)$ equivariance preserves relative angles between the inputs. 

Formally, $\varphi^{l}$ can be thought of as a Goldstone mode corresponding to the spontaneous breaking of $U(1)_{a} \times U(1)_{b}$ to a diagonal subgroup, as we explain in more detail in \eqref{phi_is_a_goldstone}. In physics, time-evolution of Goldstone modes is highly constrained and the modes generally evolve slowly. In our setting the expression \eqref{phievolve} can be thought of as an extreme version of this, where the mode turns out to be completely independent of time. 

The evolution equation for the magnitude $\Delta$ is
\be
\setlength{\abovedisplayskip}{3pt}
\setlength{\belowdisplayskip}{3pt}
\Delta^{l+1} = \frac{4(c^2 - \Delta^2)}{c^2} \int du_1~ du_2~ u_1 u_2 \exp\le(-(u_1^2 + u_2^2)\ri) I_1\le(2u_1 u_2 \frac{\Delta}{c}\ri) \phi(\frac{u_1}{\al}) \phi(\frac{u_2}{\al}) \label{evoleq-app} 
\ee
where to lighten the notation we do not make explicit that all quantities on the right-hand side are evaluated at $l$, and where $\al \equiv \frac{\sqrt{c}}{\sig_{w}\sqrt{c^2-\Delta^2}}$. Given a solution for $c^{l}$ this can be explicitly solved for $\Delta^{l}$, and we can search for fixed points by searching for a solution with $\Delta^{l+1} = \Delta^{l} = \Delta^{\star}$. 

\begin{figure}
    \centering
    \begin{subfigure}[t]{0.49\linewidth}
        \centering
        \includegraphics[width=\linewidth]{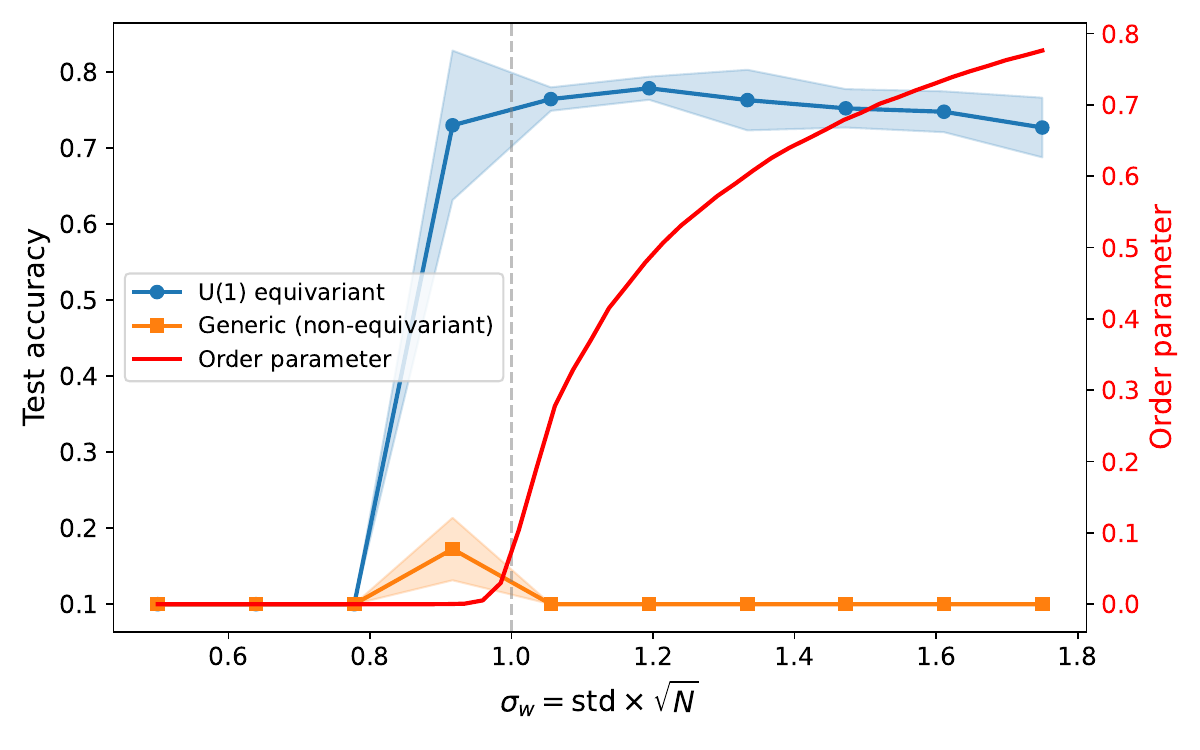}
        %\caption{$U(1)$ equivariant fully connected model with 64 features and 100 layers. Note that the model with no equivariance does not train at all, except slightly in the vicinity of the phase transition (the usual ``edge of chaos'').}
        \label{fig:test-accuracy-phasetransition-u1}
    \end{subfigure}
    \hfill
    \begin{subfigure}[t]{0.49\linewidth}
        \centering
        \includegraphics[width=\linewidth]{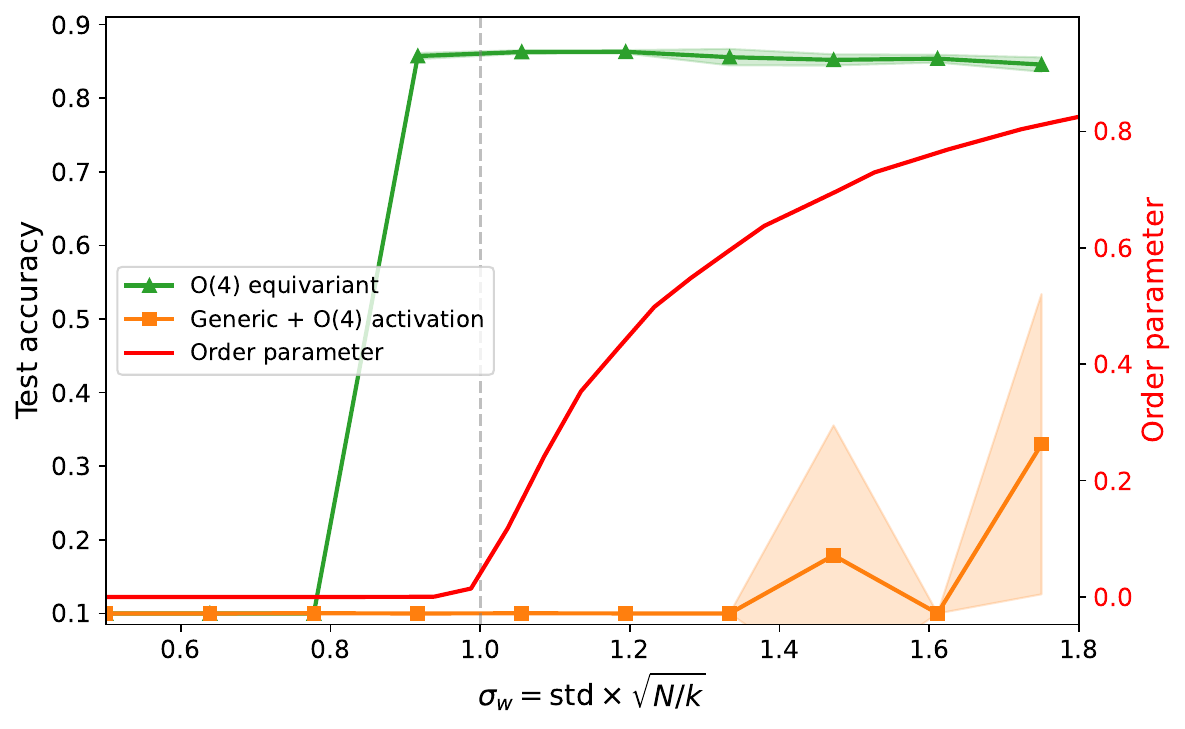}
        %\caption{$O(4)$ equivariant fully connected model with 64 features and 100 layers.}
        \label{fig:test-accuracy-phasetransition-o4}
    \end{subfigure}
    \vspace{-5mm}
    \caption{Test accuracy after 5 epochs on Fashion-MNIST for $U(1)$ (left) and $O(4)$ equivariant (right) models with 64 features and 100 layers. We compare against models with the same nonlinearity but generic linear layers, thus isolating the effect of the equivariance. We vary $\sig_{w}$ to probe the symmetry breaking phase transition, normalizing it so that the transition is always at $\sig_{w} = 1$. The training of the equivariant models appears to begin precisely when we enter the SSB phase, as measured by the order parameter $c^{\star}$, which we compute at initialization from \eqref{cstardef} averaged over the last 30 layers. Training of the non-equivariant models is poor, as expected for deep networks with no mitigation.}
    \vspace{-5mm}
    \label{fig:test-accuracy-phasetransition}
\end{figure}

Interestingly, we do not find any non-trivial fixed points in $\Delta$; on both sides of the phase transition of $\sig_{W} = 1$ we find that the only stable fixed-point is $\Delta = 0$. At large $l$ $\Delta$ approaches $0$ as $\Delta^{l} \sim \exp\le(-\frac{l}{\xi_{\Delta}}\ri)$, where the depth scale $\xi_{\Delta}$ can be computed from these expressions. In experiments we find that this depth scale is large enough that $\Delta^{l}$ remains nonzero over many tens of layers.

Although the magnitude $\Delta^{l}$ decays with depth, it does so sufficiently slowly to preserve relative phase information over many layers, while the phase itself remains exactly conserved due to \eqref{phievolve}. We thus believe that correlation information is propagated by the phase degree of freedom. \footnote{It does not seem impossible that a refinement of the dynamics could result in true non-trivial fixed point for $\Delta$ as $l \to \infty$; such a model would presumably have even better depth-generalization capacities, and is an interesting topic for future study.}  

\if0
\begin{table}[t]
\centering
\small
\renewcommand{\arraystretch}{1.25}
\begin{tabularx}{\textwidth}{
  >{\raggedright\arraybackslash}p{0.18\textwidth}
  >{\raggedright\arraybackslash}p{0.23\textwidth}
  >{\raggedright\arraybackslash}p{0.23\textwidth}
  >{\raggedright\arraybackslash}p{0.23\textwidth}
}
\hline
\textbf{Model (Theory)}
& \textbf{Before critical point}
& \textbf{At critical point}
& \textbf{After critical point} \\
\hline

Generic (DIP)
& Ordered phase; inputs collapse to a fixed point, so \textbf{information is lost}.
& Edge of chaos; long-range propagation is possible.
& Chaotic phase; input correlations decorrelate, so \textbf{information is lost}. \\

Equivariant (SSB / Goldstone)
& Unbroken phase; $c^\ell \to 0$, so \textbf{information is lost}.
& Symmetry-breaking transition; $c^\star$ starts to become nonzero.
& SSB phase; $c^\star > 0$, so \textbf{symmetry-direction information is protected}. \\

\hline
\end{tabularx}
\caption{
Comparison between deep information propagation (DIP) and the equivariant Goldstone-mode mechanism.
}
\label{tab:dip_vs_equivariant_ssb}
\end{table}
\fi

\subsection{Empirical study of training dynamics}

% \looseness=-1
To summarize, we expect that very deep neural networks with $U(1)$ equivariance should train better provided we are in the symmetry spontaneously broken phase, i.e. that $\sig_{W} > 1$. Empirically, this is precisely what we see in Figure \ref{fig:test-accuracy-phasetransition}. Unlike classical signal propagation theory, which requires fine-tuning to a critical point (the edge of chaos), here stable trainability emerges over a wide range of parameters provided we are in the symmetry-broken phase, see Figure \ref{fig:generic_eoc} in the Appendix. %for a further illustration. 

% \begin{wrapfigure}{r}{0.5\linewidth}
%     \centering
%     \vspace{-1mm}
%     \includegraphics[width=\linewidth]{mnist_layer_scan_fmnist_paper_notitle.pdf}
%     \caption{Demonstration of how performance on Fashion-MNIST degrades with increasing layer number for a generic network and for different sorts of equivariance. Note the $O(4)$ model does not degrade at all.}
%     \label{fig:fmnist_with_layers}
%     \vspace{-10mm}
% \end{wrapfigure}

\begin{wrapfigure}{r}{0.5\linewidth}
    \centering
    \vspace{-4mm}
    \includegraphics[width=\linewidth]{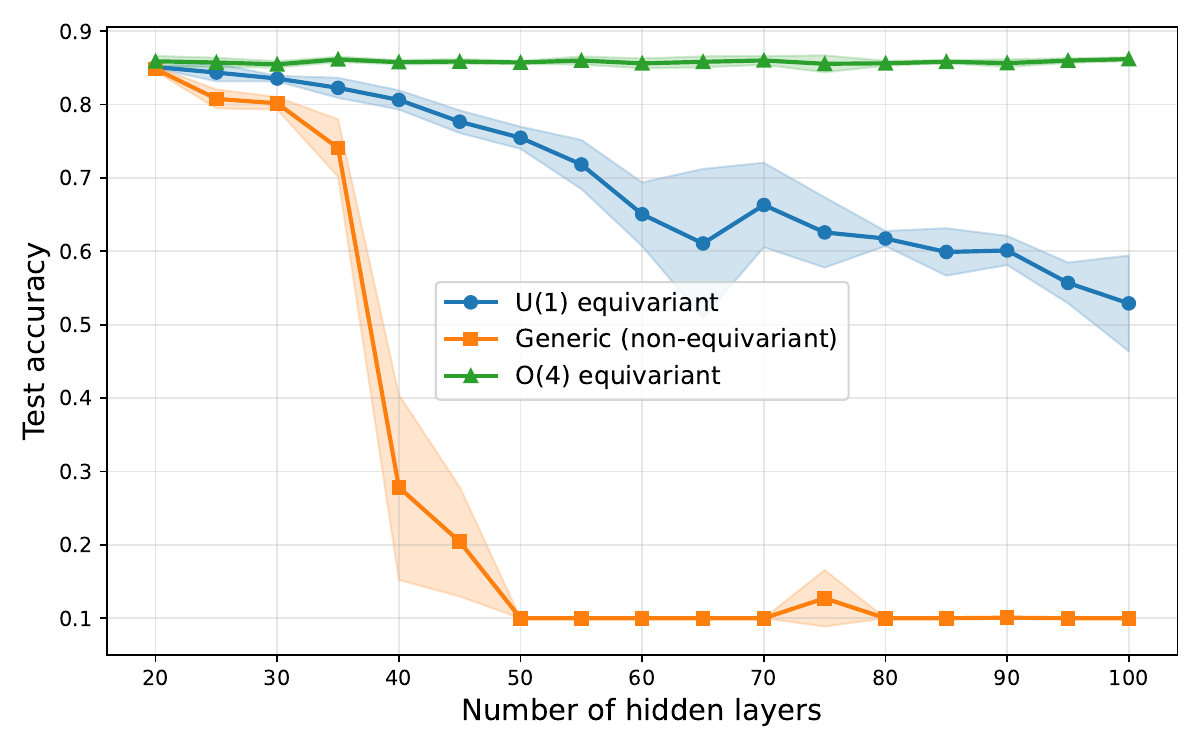}
    \caption{Demonstration of how performance on Fashion-MNIST degrades with increasing layer number for a generic network and for different sorts of equivariance. Note the $O(4)$ model does not degrade at all.}
    \label{fig:fmnist_with_layers}
    \vspace{-5mm}
\end{wrapfigure}

We also consider enlarging the equivariance to larger groups such as $O(k)$: such larger groups allow more information to be stored in the Goldstone modes, and we see that they correspondingly have stronger large-depth performance, as shown in Figure \ref{fig:fmnist_with_layers}.

It is interesting to further investigate the mechanism behind this improved performance. We believe that part of this arises from the coherent phase $\varphi^{l}$ explained in subsection \ref{sec:large_N}. It can also be understood in terms of the flow of gradients. In particular, we noted in \eqref{protected_jac} that a certain component of the Jacobian of the input-output map is {\it protected}, in that it is related to an order parameter which is nonzero in the SSB phase. %This protected component can be interpreted as arising from the neural analogue of a Goldstone mode, providing a stable, low-energy channel through which signals and gradients can propagate across depth.

\begin{figure}
\vspace{-8mm}
    \centering
    \begin{subfigure}[t]{0.48\linewidth}
        \centering
        \includegraphics[width=\linewidth]{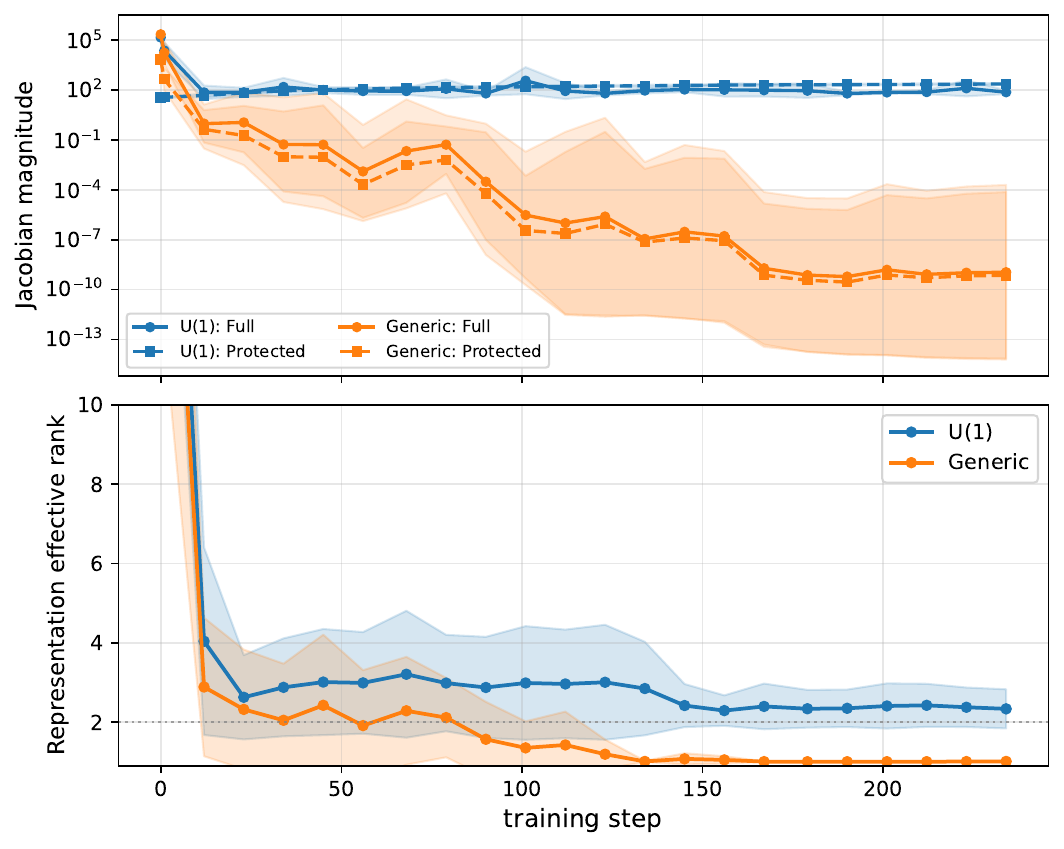}
        \caption{In top panel, we plot the norm of the input/output Jacobian and its protected component \eqref{protected_jac} for $U(1)$ equivariant and generic models. In bottom panel, we show the effective rank of the representation at the second-to-last layer of the network along training for 1 epoch.}
        \label{fig:jacobian_ref}
    \end{subfigure}
    \hfill
    \begin{subfigure}[t]{0.48\linewidth}
        \centering
        \includegraphics[width=\linewidth]{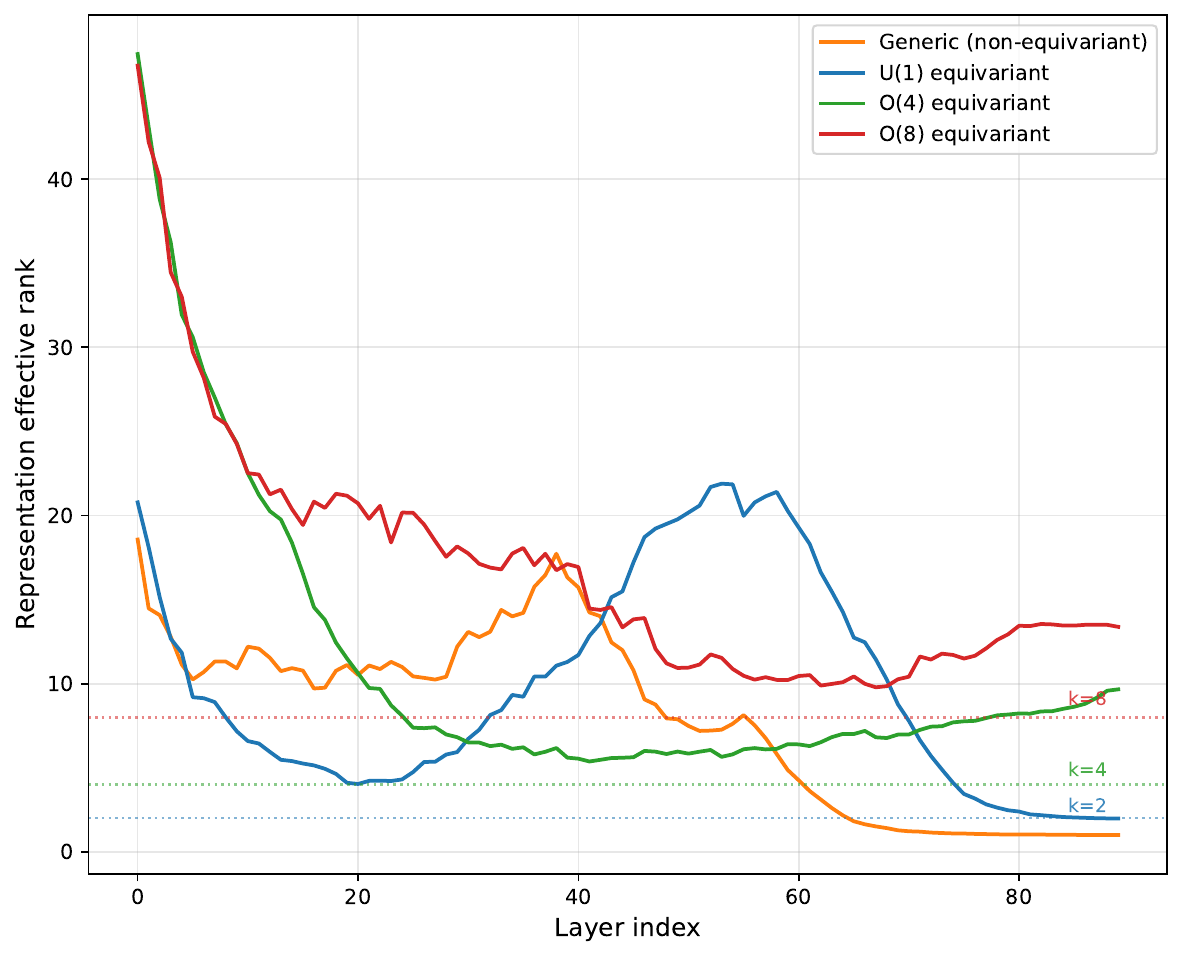}
        \caption{How effective rank of representation changes with layer for a very deep network trained on Fashion-MNIST. The generic network can collapse completely, but for equivariant networks the minimum possible rank is $k$ for $O(k)$.}
        \label{fig:rep_rank}
    \end{subfigure}
    \caption{Effective rank dynamics during training (left) and across network depth (right).}
    \vspace{-5mm}
    \label{fig:rank_combined}
\end{figure}

In Figure \ref{fig:jacobian_ref} we plot the norm of the full Jacobian and its protected component along training for equivariant and generic models on Fashion-MNIST. Note that the Jacobian for the generic model collapses as we train; this appears to be related to the rank-collapse phenomenon reported in \cite{daneshmand2020batch} when training a deep generic network without mitigations. However, the Jacobian for the equivariant model remains healthy: the protected component {\it cannot} vanish, and this acts as a clean channel for information to flow. Relatedly, the rank of the representation in the equivariant model cannot vanish, as the minimum dimension of a representation on which the $U(1)$ symmetry can act nontrivially is $2$. In Figure \ref{fig:rep_rank} we study this effective rank as we move through the layers for various symmetry groups. For $O(k)$ the dimension $k$ of the fundamental representation always provides a floor on the effective rank of the representation. Thus we see that equivariance provides a new and very simple way to ensure representation diversity in deep networks.  

%It is also interesting to consider the {effective \it rank} of the representations as we progress through the network, viewed as a function of the distribution of the singular values of the matrix at each layer in (batch, feature) space. It was shown in \cite{daneshmand2020batch} that generic deep networks tend to compress representations into one-dimensional subspaces, preventing training. In Figure \ref{fig:rep_rank} we see that our equivariant networks alleviate this somewhat: indeed the lowest dimensional linear subspace on which $O(k)$ can act faithfully has dimension $k+1$, and we see that this acts as an effective floor on how much the internal representations can collapse in an equivariant network.

To conclude, in this section we have demonstrated an analytic understanding of the phase diagram within the simplest possible context of an MLP with $U(1)$ equivariance, and showed that when the symmetry is spontaneously broken we obtain a clear advantage in training. Interestingly, in the SSB phase, even away from the critical point it is possible to train a 100-layer network with no other mitigations such as LayerNorm, skip connections, etc.

\section{Experimental results} 
\label{sec:experiments}
To study the practical implications of the information propagation capabilities afforded by spontaneously broken symmetries, we now consider tasks where significantly `deeper' long-range correlations are natural. Specifically, we consider two sequence modeling benchmarks from the literature where long-time dependency modeling is critical: a simple synthetic long-delay copying task \cite{hochreiter1997long, graves2014neural, henaff2016recurrent, gu2020improving}, and a more complex sequential image classification task (permuted sequential MNIST) \cite{le2015simple, arjovsky2016unitary, keller2023traveling}. Similar to the above Fashion-MNIST experiments, we compare generic models to equivariant counterparts, but now reconsider the `depth' over which information must propagate as recurrent iterations. % Similar to the above Fashion-MNIST experiments, we compare generic non-equivariant models to both $U(1)$ equivariant and $O(k)$ equivariant models, but now reconsider the `depth' over which information must propagate as recurrent iterations.  % Similarly, we study graph neural networks in regimes where information must be propagated over long distances in the graph and demonstrate that spontaneously broken symmetries significantly improve the networks' ability to accomplish this as well.   

\paragraph{Models, Training and Evaluation.} In the following, we use simple Recurrent Neural Networks (RNN) \cite{elman1990finding} and Gated Recurrent Unit networks (GRU) \cite{cho2014learning} as our baseline architectures. To construct the equivariant counterparts, we follow the same program as outlined in Section \ref{sec:model_theory}. % converting all recurrent operations to be equivariant with respect to the specified symmetry group, with special input and output layers that map the real valued input to the symmetry group space and back to the output space respectively. 
%
% Specifically, let $\mathbf{h}^t$ denote the hidden state at time $t$ and $\mathbf{u}^t = \mathbf{E} \mathbf{x}^t$ be the learned input embedding into the corresponding equivariant `capsule' space (e.g. via increasing dimension followed by reshaping into capsules).
Specifically, let $\mathbf{h}^t$ denote the hidden state, with $N$ counting capsules ($U(1)$: has $2N$ real coordinates; $O(k)$: has $Nk$), and let $\mathbf{u}^t = \mathbf{E} \mathbf{x}^t$ embed real inputs into this capsule space (e.g. via a learned map and reshaping).
With $\mathbf{h}^0=\mathbf{0}$, the equivariant RNNs for symmetry group $G \in \{U(1), O(k)\}$ are defined as 
\begin{equation}
\setlength{\abovedisplayskip}{3pt}
\setlength{\belowdisplayskip}{3pt}
\mathbf{h}^{t} = \phi_G\!\left(\mathbf{u}^{t} + \mathbf{W}_G \mathbf{h}^{t-1}\right), \quad \text{with} \quad
\mathbf{W}_{U(1)}\in\mathbb{C}^{N\times N},
\quad   
(\mathbf{W}_{O(k)}\mathbf{h}^t)_{\alpha a}=\sum_{\beta} \mathbf{A}_{\alpha\beta} \mathbf{h}^t_{\beta a},
    \label{eq:equiv_rnn}
\end{equation}
and $\mathbf{A} \in\mathbb{R}^{N \times N}$. 
For $G=U(1)$, $\mathbf{h}^t\in \mathbb{C}^N$ and $\phi_G$ is the radial nonlinearity from \eqref{nonl}; for $G=O(k)$, 
$\mathbf{h}^{t}\in \mathbb{R}^{N \times k}$ has an additional $k$ internal `capsule' dimensions, $\phi_G$ acts on the $O(k)$-invariant radius of each block, and the recurrent map mixes capsules but acts isotropically within each capsule, i.e. the recurrent map has the form $\mathbf{A} \otimes I_k$, as detailed in Appendix~\ref{app:equivariant-layers}. 
For the $U(1)$-equivariant GRU, the gates are real, phase-invariant functions of the previous hidden state:
\begin{equation}
\begin{aligned}
\tilde{\mathbf h}^{t} &= \phi_{U(1)}\!\left(\mathbf u^{t} + \mathbf W_h(\mathbf r^{t} \odot \mathbf h^{t-1})\right), \qquad
& \mathbf r^{t} = \sigma\!\left(\mathbf W_{xr}\mathbf x^{t} + \mathbf W_{hr}|\mathbf h^{t-1}| + \mathbf b_r\right),\\
\mathbf h^{t} &= \mathbf z^{t}\odot \mathbf h^{t-1} + (1-\mathbf z^{t})\odot \tilde{\mathbf h}^{t}, \qquad
& \mathbf z^{t} = \sigma\!\left(\mathbf W_{xz}\mathbf x^{t} + \mathbf W_{hz}|\mathbf h^{t-1}| + \mathbf b_z\right).
\end{aligned}
\label{eq:u1_gru}
\end{equation}
\looseness=-1
Here $|\mathbf{h}|$ is taken elementwise, so the gates are invariant under the global $U(1)$ phase rotation, while the candidate hidden update remains equivariant. For readouts, we flatten the hidden state capsule dimensions and perform a linear map to the output space.
For all models, we sweep over weight initialization schemes (identity vs. uniform random), learning rates $\lambda \in \{10^{-3}, 10^{-4}, 10^{-5}\}$, and activation functions $\{\tanh, \mathrm{ReLU}\}$ (for the RNNs). Model selection is performed on a validation set, and all results are reported on the test set. Parameter counts are reported in real trainable parameters, where complex weights count as two real parameters. Full details are provided in Appendix \ref{app:experiment_details}.

\begin{figure}[t]
\vspace{-12mm}
\centering
        \includegraphics[width=\linewidth]{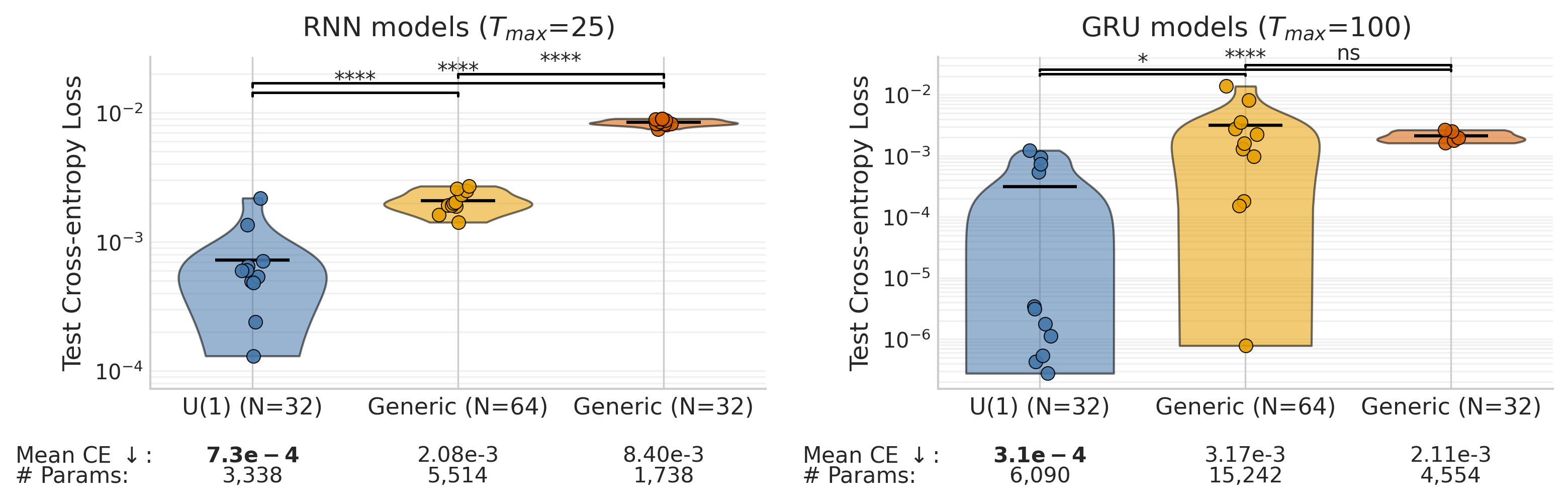}
        \label{fig:test-accuracy-phasetransition-o4-p}
    \vspace{-5mm}
        \caption{Comparison of RNNs and GRUs with $U(1)$ equivariant counterparts on copy tasks of max delay $T_{max}=25$ and $T_{max}=100$ respectively. We see that the $U(1)$ models significantly outperform the non-equivariant counterparts for both architectures, even when generic baselines have more real trainable params.}    
        \vspace{-3mm}
    \label{fig:copy_joint}
\end{figure}

% \subsection{Recurrent Neural Networks}
\paragraph{Variable Delay Copy Task.}

% \begin{wrapfigure}{r}{0.4\linewidth}
%     \centering
%     \vspace{-7mm}
%     \includegraphics[width=0.99\linewidth]{neurips/copy_T100_plots_rnn_only.png}
%     \caption{Comparison of RNNs on the copy task with $T_{max}=25$; U(1) RNNs dramatically outperform the non-equivariant counterparts, even those double the size.}
%     \vspace{-10mm}
%     \label{fig:copy_T25}
% \end{wrapfigure}
\looseness=-1
We first study a variable-delay version of the standard copy task \citep{hochreiter1997long, graves2014neural, henaff2016recurrent, gu2020improving}. In each example, the network is presented with a length-10 sequence of one-hot data tokens sampled from 8 data symbols, with two additional symbols reserved for `blank' and `go'.
% a random 10-dimensional one-hot sequence of length 10 to memorize, followed by a sequence of $T$ delay tokens. 
After the delay period, the special `go' token is provided, indicating that the network should reproduce the original input sequence. Unlike the standard copy task, which uses a fixed delay, we sample the delay
independently for each example as $T \sim \mathrm{Unif}\{0,\ldots,T_{\max}\}$. This prevents models from relying on a fixed output time, requiring the `go' token to trigger readout. The target sequence is blank at all timesteps except for the $10$ positions following the `go' token, where the model must output the initially presented sequence. This task therefore tests the long-range dependency modeling capabilities of a sequence model over variable sequence lengths.
% 
% \begin{wrapfigure}{r}{0.4\linewidth}
%     \centering
%     \vspace{-0mm}
%     \includegraphics[width=0.99\linewidth]{neurips/copy_T100_plots_gru_only.png}
%     \caption{Comparison of GRUs and U(1) GRUs on the variable-delay copy task with $T_{max}=100$. Significantly more U(1) GRU Models solve the task, and require significantly fewer parameters (5K) than comparable baseline GRUs (15K).}
%     \vspace{-12mm}
%     \label{fig:copy_T100}
% \end{wrapfigure}
% In Figure \ref{fig:copy_joint} (left), we compare U(1) equivariant simple RNNs with $N=32$ hidden units against non-equivariant counterparts with both 32 and 64 hidden units for $T_{max}=25$. As can be seen, the equivariant model dramatically outperforms both an RNN of the same size, and one with more than double the parameters across 10 random seeds.
In Figure \ref{fig:copy_joint} (left), we compare $U(1)$ equivariant RNNs with $N=32$ complex capsules (64 real coordinates) against non-equivariant RNNs with 32 or 64 real hidden units for $T_{\max}=25$. The equivariant model dramatically outperforms both baselines across 10 random seeds, including the 64-unit baseline with comparable real hidden dimensionality. 
% In Figure \ref{fig:copy_joint} (right), we compare U(1) equivariant GRUs with 32 hidden units on a longer version of the same task ($T_{max}=100$) with comparable non-equivariant GRUs with both 32 and 64 hidden units. Across 10 random seeds, we again see that the U(1) equivariant GRUs solve the task significantly more frequently than non-equivariant counterparts, which we attribute to more stable training, and thereby achieve lower mean loss with significantly fewer parameters (5K vs. 15K).
In Figure 6 (right), we compare $U(1)$ equivariant GRUs with $N=32$ complex capsules on a longer task ($T_{\max}=100$) against non-equivariant GRUs with 32 or 64 real hidden units. Across 10 random seeds, the $U(1)$ GRUs solve the task more often, suggesting more stable training, and achieve lower mean loss with fewer real scalar parameters (6K vs. 15k). 
On $T_{\max}=200$, we again find that with the same hyperparameter sweep, the only model to have a seed which solves the task is the $U(1)$ GRU, achieving a test cross-entropy loss of $3.9 \times 10^{-7}$.

\begin{wrapfigure}{r}{0.4\linewidth}    
    \centering
    \vspace{-5mm}
\includegraphics[width=0.94\linewidth]{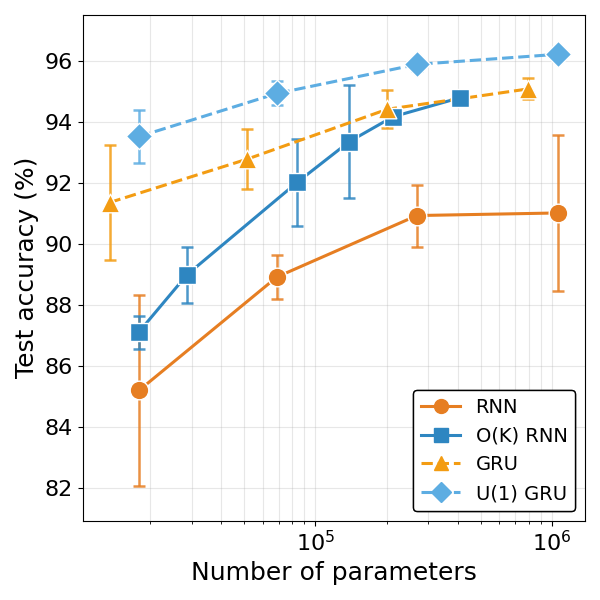}
    \vspace{-2mm}
    \caption{Test accuracy vs. \# real trainable params on psMNIST. Equivariant RNNs and GRUs consistently outperform non-equivariant models at all parameter ranges.}
    \vspace{-4mm}
    \label{fig:psmnist}
\end{wrapfigure}
\paragraph{Sequential Image Classification.} We next compare models on a pixel-by-pixel image classification task that requires both long-sequence memory and simultaneous processing of the information. Specifically, we study a variant of sequential MNIST \cite{le2015simple}, where each pixel of a $28\times28$ grayscale MNIST image is presented to the network one-at-a-time, and the task of the network is to predict the class of the handwritten digit after all 784 steps from its final hidden state. 
To make the task more challenging, following prior work \cite{arjovsky2016unitary}, we perform a fixed random permutation of the pixel order for both training and testing. This permutation reduces the short-range correlations that are typically present in standard sequential MNIST, which can reduce the effective length of the task. As seen in Figure \ref{fig:psmnist}, the equivariant RNNs and GRUs consistently outperform the non-equivariant models across all ranges of parameters (as a function of hidden state size). Perhaps most impressively, we see that an $O(k)$ equivariant simple RNN ($k=24$) can reach the performance of a comparable GRU despite having a simpler computational graph with no internal gating. While not state-of-the-art, these models perform comparably to specifically designed long-sequence models such as IndRNN \cite{li2018independently}, Lipschitz RNN \cite{erichson2021lipschitz}, coRNN \cite{cornn} and LEM \cite{LEM}, all of which reach between 96\% to 97\% accuracy on the same task. As can be seen by the performance boost for the GRU, we speculate that the spontaneous symmetry breaking mechanism may be compatible with some of these architectures, and therefore may further improve performance when combined with such models. We leave further exploration of such joint models to future work and present these results as preliminary evidence for the practical performance boost that spontaneous symmetry breaking may yield in a precisely controlled comparative setting. % \NI{should we comment on how these things compare to the state of the art? there exist still better models, is that right?} 

% \begin{table}[h]
%   \centering
%   \label{Copy}
%   \caption{Permuted Sequential MNIST}
%   \begin{tabular}{l l l | c c }
%     \toprule
%                  &  N & K & Test Accuracy & \# $\theta$ \\ \hline \hline
%         $U(1)$ RNN & 256 & - & - & - \\
%         $U(1)^K$ RNN & 256 & - & - & -\\ 
%         \hline
%         $O(K)$ RNN & 256 & 6 & 92.14 & 84k\\
%         $O(K)$ RNN & 256 & 12 & 91.06 & 102k\\
%         $O(K)$ RNN & 256 & 24 & 94.14 & 139k\\
%         $O(K)$ RNN & 256 & 48 & \textbf{94.22} & 213k\\ 
%         \hline
%         Simple RNN & 256 & - & - & - \\
%         Simple RNN & 512 & - & 89.44 & 269k \\
%         Simple RNN & 1024 & - & 90.76 & 1.06M \\
%         Simple RNN & 1536 & - & 91.13 & 2.38M \\
%         Simple RNN & 2048 & - & 93.31 & 4.22M \\
%     \bottomrule
%   \end{tabular}
% \end{table}

%\subsection{Graph neural networks}
%\NI{If Nabil can get some results here; I think for now I will not focus on this, and leave it instead for the rebuttal or future versions.} 
\section{Conclusion}

In this work we have demonstrated that an analogue of Goldstone modes can lead to concrete benefits in information storage and training in deep networks. This leads to concrete benefits for RNNs, and interestingly allows for training very deep networks without the usual architectural heuristics such as LayerNorm, BatchNorm, skip connections, etc. 

\paragraph{Topological defects.} We note that in physics spontaneously broken symmetries are associated with two sets of phenomena which control the long-time physics: Goldstone modes (whose deep-learning analogue we have extensively studied here), and {\it topological defects}, localized objects such as vortices, domain walls and hedgehogs (monopoles) which persist stably in time \cite{coleman1988aspects,rajaraman1982solitons}. The latter exist even in the case of spontaneously broken discrete symmetries such as $\mathbb{Z}_2$. They provide a logically distinct route towards stable information propagation through layers of a network. Interestingly, there are some indications that such structures may play a role in neuroscience \cite{xu2023interacting}. They remain relatively unexplored in a deep learning context (though see \cite{iqbal2025topological} for some preliminary investigations).  In Appendix \ref{app:topo_defects}, Figures \ref{fig:vortices_0} and \ref{fig:vortices_1} we present a preliminary study of such topological defects in a 2D convolutional RNN (similar to Recurrent CNNs \cite{pinheiro2014recurrent, liang2015recurrent} and ConvLSTMs \cite{shi2015convlstm, ballas2016delving, wang2017predrnn, pmlr-v202-keller23a} used for spatiotemporal forecasting) 
% \NI{A bit worried about whether the domain wall is an artifact (see email). Also: Andy is there a canonical reference for "2d convolutional RNN", is this completely standard?} 
augmented with $U(1)$ symmetry, where we see long-lived vortices emerging when solving the copy task, though their precise role remains to be understood.  

We conclude by noting that the ideas here provide a new use of equivariance in neural networks -- which was previously restricted to symmetries associated with the task or data distribution -- and we anticipate further applications in the future.

\looseness=-1
\paragraph{Limitations.} While our current experiments are limited to simple benchmarks, they demonstrate the theory can apply beyond idealized analyses, including to trained neural networks in standard supervised learning settings, with the potential to improve performance in large-depth regimes.

\section*{Acknowledgments and disclosure of funding} This work was supported by a grant from the
Simons Foundation (PD-Pivot Fellow-00004147, NI). NI is supported in part by the STFC under grant number ST/T000708/1. This work has been made possible in part by a gift from the Chan Zuckerberg Initiative Foundation to establish the Kempner Institute for the Study of Natural and Artificial Intelligence at Harvard University: TAK is supported by the Kempner Institute Research Fellowship. 
TM is supported by the ERC Starting Grant LEGO-3D (850533).
TM is also supported by the Google PhD Fellowship. TM acknowledges his affiliation with the ELLIS (European Laboratory for Learning and Intelligent Systems) PhD program.

\bibliographystyle{utphys}
\bibliography{all}

%%%%%%%%%%%%%%%%%%%%%%%%%%%%%%%%%%%%%%%%%%%%%%%%%%%%%%%%%%%%
\newpage
\appendix
\begin{appendices}
\addcontentsline{toc}{section}{Appendix} % Add the appendix text to the document TOC
\vspace{-20mm}
\part{} % Start the appendix part
\parttoc % Insert the appendix TOC

\section{Background on Goldstone modes} \label{app:goldstone} 
To make this work self contained we  provide a brief review of Goldstone modes in physics. This material is standard, see e.g. \cite{Peskin:1995ev} for a textbook treatment. 

Let us imagine an interacting {\it complex scalar field} $\phi$, i.e. a field $\phi(t,x)$ which depends on space and time through an equation of the following form:
\be
\le(\p_t^2 - \p_{x}^2\ri)\phi = -\frac{dV}{d\phi^{\dagger}} \label{eom} 
\ee
where we take the {\it potential energy} $V = r |\phi|^2 + \lam |\phi|^{4}$. This system has a $U(1)$ symmetry: e.g. given a solution to the equations of motion \eqref{eom}, one can find a new solution by rotating $\phi$ by a constant phase: $\phi(t,x) \to e^{i\al} \phi(t,x)$. 

\begin{figure}
    \centering
    \includegraphics[width=0.75\linewidth]{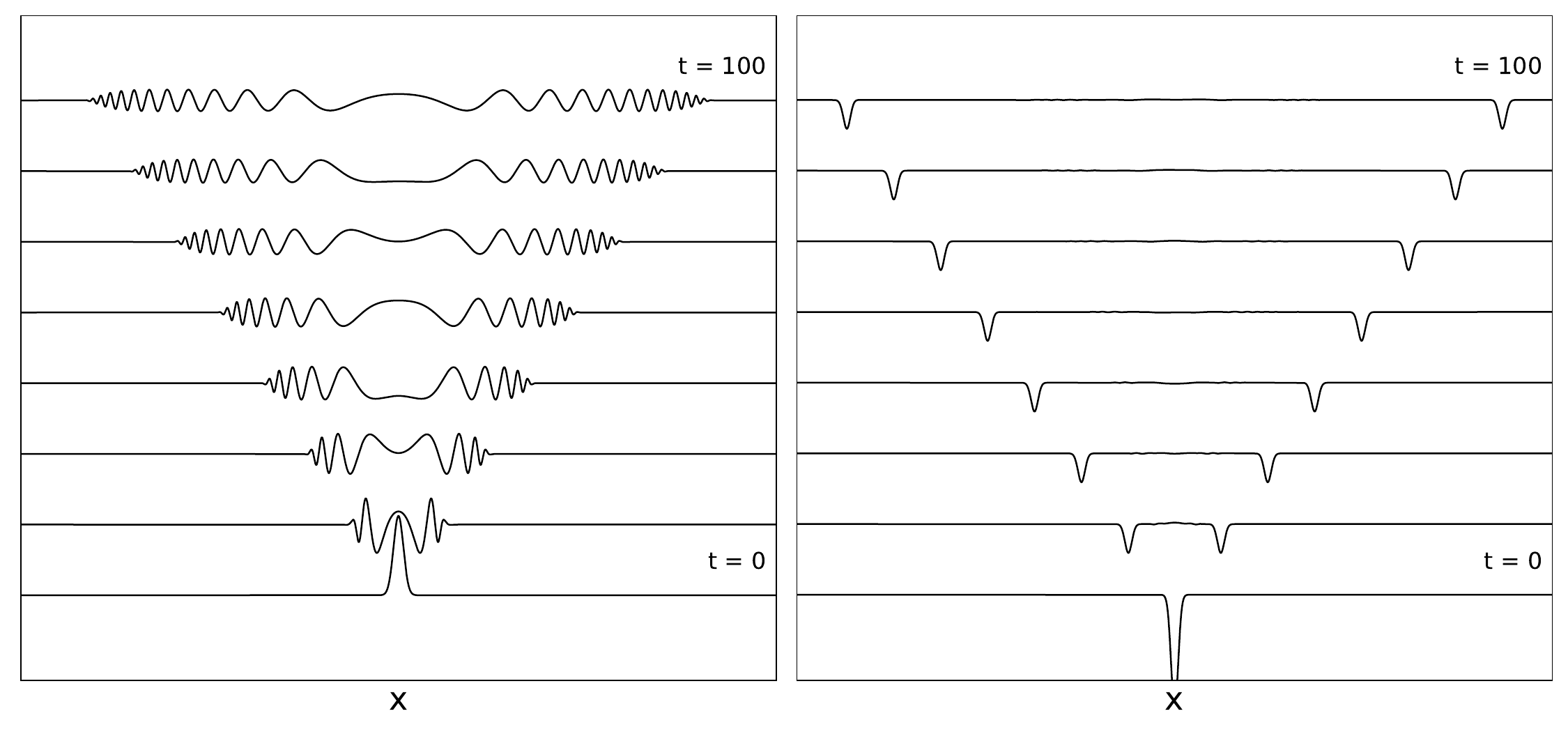}
    \caption{Excitations in symmetry unbroken (left) and symmetry spontaneously broken (right) phases respectively. We plot the time evolution of the real part of the complex scalar field described in \eqref{eom}. On left the wave-packet spreads apart and degrades; on the right we see the Goldstone mode propagating coherently across long distances.}
    \label{fig:goldstone}
\end{figure}

Now the dynamical behavior of this field depends sensitively on the sign of the parameter $r$, as we show in Figure \ref{fig:potential}. If $r > 0$, then the configuration on which $V$ is minimized (i.e. the {\it equilibrium} configuration) is $\phi = 0$. Note that this equilibrium is {\it invariant} under the $U(1)$ symmetry; for this reason the system with $r > 0$ is sometimes called the phase with {\bf unbroken symmetry}. One can study small excitations around this equilibrium. If we consider a small perturbation with spacetime dependence $e^{-i\om t + i k x}$, then by linearizing the equations of motion about $\phi = 0$ and plugging in the above ansatz, we can show that the oscillation frequency of the wave behaves as:
\be
\om = \sqrt{k^2 + r} \ .
\ee
which means that even a very long wavelength wave has a finite oscillation frequency $\om \sim \sqrt{r}$. Relatedly, this means that wave packets will eventually disperse and dissipate over long distances, as one can see in the left panel of Figure \ref{fig:goldstone}. 

We can also consider the system with $r < 0$. Now the lowest energy configuration is no longer unique: the energy is minimized on a {\it ring}, i.e. everywhere that $|\phi| = \sqrt{\frac{-r}{2\lam}}$, as we show in the right panel of Figure \ref{fig:potential}. All points along this ring -- related by a $U(1)$ rotation -- have the same energy, and thus the system will simply choose one of them: this is called {\bf spontaneous symmetry breaking}. We can again now consider small perturbations around this equilibrium. However low energy fluctuations now only modulate the phase of the field, i.e. $\phi(x,t) = \sqrt{\frac{-r}{2\lam}} e^{i\th(x,t)}$. Again finding the dispersion relation of small-amplitude oscillations, one finds that the frequency of the excitation now behaves as
\be
\om = |k| \label{goldstone} 
\ee
i.e. the oscillation frequency of the wave goes to zero as the wavelength becomes very long. This happens because $\om$ is related to the energy of the excitation -- but a spatially homogenous excitation with $k \to 0$ is now just a symmetry action on the equilibrium configuration, which costs no energy, and thus the frequency of the excitation must vanish. Furthermore, the linear dependence of the energy on the wavenumber means that there is no broadening of the wavefront, and the wave travels for long distances with no dissipation, as we can see in the right panel of Figure \ref{fig:goldstone}. 

This is a specific example of a very general principle, called {\bf Goldstone's theorem} \cite{Goldstone:1962es}: when a system has a symmetry which is spontaneously broken by the equilibrium configuration, then it has {\it long-lived} excitations called {\it Goldstone modes}, of which the wave in \eqref{goldstone} is an example. 

In this work we will demonstrate a similar principle at play in deep learning. In particular, we will show that when a neural network has internal layers which are equivariant under a global symmetry, there is a region in hyperparameter space (analogous to the region $r < 0$) where one can imagine that a symmetry is spontaneously broken, and that its corresponding Goldstone mode allows information to propagate coherently through the network.

\section{Implementation of equivariant layers} \label{app:equivariant-layers}

Here we describe how equivariant layers such as those described in \eqref{equivlayers} are implemented. 

\paragraph{$\mathbf{O(k)}$:} We begin by studying equivariance under the more general $O(k)$ symmetry. We consider a representation $\bf{x} \in \mathbb{R}^{N \times k}$. We consider acting on the $k$ index with the $O(k)$ symmetry, i.e. letting $\alpha,\beta$ run over $N$ blocks and $a,b$ run over the $k$ components of the block, the symmetry acts as
\be
x'_{\al a} = O_{ab} x_{\al b} \label{Okdef} 
\ee
where $O_{ab} \in O(k)$ is an $O(k)$ rotation, i.e. it rotates the components of each block within itself but does not mix the blocks. We would now like to build a neural network layer which is equivariant under the symmetry. 

This is done by considering a weight matrix $W$ which takes the form
\be
W_{\al a;\beta b} = w_{\al\beta} \delta_{ab} \label{Wdef} 
\ee
i.e. it acts as the identity within each block but mixes the blocks together in an unconstrained manner. This clearly commutes with the general $O(k)$ rotation in \eqref{Okdef}; by Schur's lemma it is also the most general such matrix provided $O(k)$ is non-Abelian, i.e. that $k>2$. Note that we have only $N^2$ free parameters, labeled by $w_{\al\beta}$, as opposed to the $(Nk)^2$ we could have had for the non-equivariant model. 

We also consider a non-linearity which acts on the $O(k)$-invariant radius of each block, i.e.
\be
\phi(x)_{\al a} = \frac{\tanh(|x_{\al}|)}{|x_{\al}|} x_{\al a} 
\ee
where $|x_{\al}| \equiv \sqrt{\sum_{a=1}^k x_{\al a}^2}$, i.e. it replaces the radius of each block with its hyperbolic tangent while preserving its $O(k)$ orientation. Clearly one can replace $\tanh$ with any nonlinear function of one variable while preserving the equivariance properties. Generally one $O(k)$ layer consists of a single nonlinearity and a single linear layer. 

\paragraph{$\mathbf{SO(2) = U(1)}$:} The case of $SO(2)$ (or equivalently $U(1)$) is different. In this case the components $a,b$ run over $1,2$ and the most general choice for \eqref{Wdef} is
\be
W_{\al a;\beta b} = (w_{\al\beta} \delta_{ab} + u_{\al\beta} \ep_{ab}) \label{U1wdef} 
\ee
where now $\ep_{ab}$ is the fully antisymmetric two dimensional matrix and we now have two sets of trainable parameters $w_{\al\beta}$ and $u_{\al\beta}$. (Not if we considered $O(2)$ instead then the term in $\ep_{ab}$ would not be allowed, analogous to the case in \eqref{Wdef} -- e.g. the term in $\ep_{ab}$ is not operation under the group operation $\mathrm{diag}(-1,1) \in O(2)$). This is equivalent after a change of basis to considering a complex representation $\mathbf{z} \in \mathbb{C}^{N}$ and allowing the most general complex $W$ acting as \eqref{feedforwardU1}. In this notation the $U(1)$ equivariance comes from the fact that the forward linear map is purely holomorphic (i.e. depending on $z$ and not $z^{\dagger}$). 
\section{Path integral for stochastic systems} \label{app:path-integral} 
Here we present details of the path integral description of information propagation through the layers of a feedforward network. Path integrals provide a convenient calculational framework for discussing stochastic evolution equations \cite{martin1973statistical}, allowing for controlled derivation of the mean-field expansion, along with departures from it. In the context of neural networks these techniques were used in the classic work \cite{sompolinsky1988chaos} and have since been revisited in the modern context in \cite{crisanti2018path,schoenholz2017correspondence}. To the best of our knowledge they have not been used to study group equivariant neural networks until this work. 

\subsection{$U(1)$ equivariant feedforward network}
We will present our main derivation for the $U(1)$ equivariant version. The generalization to other symmetry groups is mostly a change of notation. In particular, we consider a fully connected network with following update rule:
\be
z^{l+1}_{i} = \sum_{j} W^{l}_{ij} \phi(z^{l}_{j}) \label{update_app} 
\ee
where now $z \in \mathbb{C}^{N}$, $W \in \mathbb{C}^{N \times N}$, and the nonlinearity (and thus the entire network) is $U(1)$ equivariant, in that $\phi(e^{i\th} z) = e^{i\th} \phi(z)$. For concreteness we take the nonlinearity to be:
\be
\phi(z) = \frac{\tanh(|z|)}{|z|} z
\ee
which replaces the magnitude of the complex number by its hyperbolic tangent. 

The primary object of study is the {\it partition function}, which is a functional of the applied ``external fields'' $h^{l}_i$. 
\be
Z[h] = \int [dz][dW] p(z^L | z^{L-1}) p(z^{L-1} | z^{L-2}) \cdots p(z^{1} | z^{0}) \prod_{l} p(W^{l}) \prod_{l} e^{h^{l}_{i} z^{l}_i + h.c.}
\ee
Here the integral takes place over all internal representations $[dz]$ as well as the initial ensemble of the inputs, as well as over all the weights $[dW]$. The official purpose of this object is that we can differentiate it with respect to the fields $h^{l}_i$ to obtain moments of the $z^{l}_i$; each derivative of the exponential brings down a factor of the corresponding $z^{l}_i$, and we find (e.g.)
\be
\frac{\p}{\p h^{l}_i} \frac{\p}{\p h^{l,\dagger}_j} \log Z[h]\bigg|_{h=0} = \mathbb{E}[z^{l}_i (z^{l}_j)^{\dagger}]
\ee
In practice however it is usually sufficient to know that this is possible, and we do not need to carry around the $h$ fields explicitly. Instead the utility of $Z$ is that its construction in terms of integrals directly shows the correlation structure of the activations. For now we will set $h$ to zero for notational simplicity and work with $Z \equiv Z[h=0]$, and we can always restore it if needed. 

Next, we assume as usual that at initialization each entry of the $W^l_{ij}$ at each level is i.i.d. Gaussian with variance $\frac{\sig_{W}^2}{N}$. Once the $W$ are fixed the evolution of the $z^l_i$ is deterministic according to \eqref{update_app}, and we find the following iterated integral representing how each input evolves through the network. 
\be
Z = \int [dz] dW e^{-\frac{N}{\sig_{w}^2}\sum_{ij,l} W^l_{ij}W^{l\dagger}_{ij}} \prod_{i,l} \delta\le(z^{l+1}_i - \sum_{j} W^{l}_{ij} \phi(z^{l}_j)\ri)
\ee
To obtain a more useful representation, we write the delta function in terms of an auxiliary field $\hat{z}_i$ introduced at each layer, i.e.
\be
\delta\le(z^{l+1}_i - \sum_{j} W^{l}_{ij} \phi(z^{l}_j\ri) = \int d\hat{z}^l_i d\hat{z}^{l\dagger}_i \exp\le(i \hat{z}^{l+1}_i\le(z^{l+1}_i - \sum_{j} W^{l}_{ij} \phi(z^{l}_j\ri) + h.c.\ri)
\ee
(where various factors of $2\pi$ have been absorbed into the definition of the measure $[dz]$, as they will not affect our conclusions). We thus end up with the following representation 
\be
Z = \int [dz d\hat{z} dW] e^{S[z,\hat{z},W]} \qquad S = -\frac{N}{\sig_{w}^2} \sum_{ij} W^{l}_{ij} W^{l\dagger}_{ij} + i \sum_{l,i} \hat{z}^{(l+1) \dagger}_i \le(z^{l+1}_i - \sum_{j} W^{l}_{ij} \phi(z^{l}_j\ri) + h.c.
\ee
where $S$ is called the {\it effective action}.

We now note that $W_{ij}^l$ appears in the action quadratically. Thus its integral is a Gaussian and can be explicitly done, leaving us with the following effective action:
\be
S  = i \sum_{l,i} \hat{z}^{l}_i z^{l}_i - \frac{\sig_W^2}{N} \sum_{l,i,j} \hat{z}^{l+1}_i\hat{z}_i^{(l+1)\dagger} \phi(z^{l}_j) \phi(z_j^{l\dagger})
\ee
We have now taken into account the integral over the $W$'s: the resulting action has a very specific form, in that each component $z_i$ interacts with every other component through the quartic term. This pattern of interactions is exactly that which is well described by mean-field theory, and it suggests that dynamics can be cleanly written in terms of collective variables whose distributions are sharply peaked in the large $N$ limit. 

To that end, we introduce the following collective field 
\be
c^{l} = \frac{1}{N} \sum_{j} \phi(z^{l}_j) \phi(z^{l}_j) \label{cdef} 
\ee
To express the dynamics in terms of this collective field, we insert 1 into the path integral in the following form:
\be
1 = \int [dc d\hat{c}] \exp\le[\sum_{l} i N \hat{c}^{l} \le(c^{l} -  \frac{1}{N} \sum_{j} \phi(z^{l}_j) \phi(z^{l}_j))\ri)\ri]
\ee
where $\hat{c}^l$ is yet another field we have introduced. Its integral acts as a delta function which enforces the relation \eqref{cdef}; thus we may replace $\frac{1}{N} \sum_{j} \phi(z^{l}_j) \phi(z^{l}_j)$ with $c^{l}$ in the effective action, leaving us with
\be
S  = i \sum_{l} \le[\sum_{i} \hat{z}^{l}_i z^{l}_i - \sig_{W}^2 c^{l} \sum_{i}\hat{z}^{l+1}_i\hat{z}_i^{(l+1)\dagger} + i N \hat{c}^{l} \le(c^{l} -  \frac{1}{N} \sum_{j} \phi(z^{l}_j) \phi(z^{l}_j)\ri)\ri]
\ee

Now the integral over the $\hat{z}^{l}_i$ is Gaussian; performing the integrals we obtain
\be
S = \sum_{l}\le[\frac{1}{\sig_{W}^2 c_{l}}  \sum_{i} z_{i}^{l+1} (z^{l+1}_{i})^{\dagger}  + i N \hat{c}^{l}\le(c^{l} -  \frac{1}{N} \sum_{j} \phi(z^{l}_j) \phi(z^{l}_j)\ri) - N \log \sig_{W}^2 c^l \ri] \label{fullac_S} 
\ee
where the last term arises from the determinant of the Gaussian integral and where (as usual for path-integrals) we ignore constant additive shifts in the action.  

So far everything we have done has been exact. We now consider the large $N$ limit. Note that both of the terms in this action scale as $N$. This indicates that the theory becomes classical in the large $N$ limit. As usual then we can evaluate the functional integral over $[dc d\hat{c}]$ in a saddle-point approximation by varying this action and demanding that it be stationary. The equation of motion for $\hat{c}^{l}$ simply tells us that $c^{l}$ satisfies \eqref{cdef}. Thus we see that each component of the $z$'s has an identical Gaussian distribution, where the width $c_{l}$ of the distribution at layer $l+1$ is determined by the product of activations at the previous layer in \eqref{cdef}. 

This is a self-consistent evolution equation for $c^{l}$:
\be
\frac{1}{2\pi i \sig_{w}^2 c^{l}}\int dz dz^{\dagger} \exp\le(-\frac{1}{\sig_{w}^2 c^l}z^{\dagger}z\ri) |\phi(z)|^2 = c^{l+1} \label{ceqn_zz} 
\ee
It is convenient to go to polar coordinates $z = r e^{i\th}$. The integral over $\th$ can be done trivially, and we rewrite in terms of $u = \frac{r}{\sig_{W} \sqrt{c}}$ so that the equation becomes
\be
2 \int du u e^{-u^2} [\phi(u \sig_{w} \sqrt{c^l})]^2 = c^{l+1} \label{app_cevolve}
\ee
This equation is the $U(1)$-equivariant analogue of Eq (3) in \cite{schoenholz2017deep}. The implications of this evolution equation are described around \eqref{cevolve} in the main text. To understand the dynamics near the transition we specialize to the case $\phi = \tanh$ and expand the integrand side in powers of $c^{l}$ and perform the integral order by order, finding:
\be
c^{l+1} = c^{l} \sig_{w}^2 -  \frac{4(c^{l})^2 \sig_{w}^4}{3} + \sO(c_{l}^3)
\ee
First considering $c^{l}$ very small so that we may ignore the quadratic term, we see that for $\sig_{w} < 1$ there is an attractive fixed point at $c^{l} \to 0$. For $\sig_{w} > 1$ this solution is now unstable and $c^{l}$ is repelled to a nonzero value. We may search for the non-trivial fixed point near the phase transition by setting $c^{l} = c^{l+1} = c^\star$, finding:
\be
c^{\star}(\sig_{w} \to 1) \approx \frac{3}{4}(\sig_{w}^2 - 1)
\ee
i.e. a mean-field exponent. 
\subsection{Two-replica solution}
We would now like to understand how the distribution of {\it two} inputs changes as we move through the network. Denoting these inputs as $z_{i,a}$ and $z_{i,b}$ where $a, b \in \{a, b\}$, we essentially have twice as many dynamical fields -- or {\it replicas} --  with a diagonal term in the action describing their dynamics as
\be
S \supset \exp\le[i \sum_{l,i,a} \hat{z}^{l+1}_{i,a} \le(z^{l}_{i,a}(t+1) - \sum_{j} W_{ij}\phi(z^l_{j,a})\ri) + h.c.\ri]
\ee
We do not repeat the full derivation here; it essentially proceeds exactly as above, except that integrating out the $W_{ij}$'s now creates a coupling between replicas, and we should now consider the following matrix-valued collective field:
\be
C^{l}_{ab} = \frac{1}{N} \sum_{j} \phi^{\dagger}(z^{l}_{j,a}) \phi(z^{l}_{j,b}) \label{cab_def} 
\ee
The important term of the resulting two-replica action is that controlling the distribution of the $z^{i}_a$ which is a simple generalization of \eqref{fullac_S}:
\be
S = \sum_{l}\le[\frac{1}{\sig_{W}^2} (C^l)^{-1}_{ab}  \sum_{i,a,b} z_{i,a}^{l+1} (z^{l+1}_{i,b})^{\dagger}  + \cdots\ri] \label{2repac} 
\ee
It is helpful to explicitly write down the components of $C_{ab}$ as:
\be
C^{l}_{ab} = \begin{pmatrix} 
            c^{l} & \Delta^{l} e^{i\varphi^{l}} \\
            \Delta^{l} e^{-i\varphi^{l}} & c^{l} 
            \end{pmatrix}
            \ee
where the diagonal correlation is precisely the $c^{l}$ studied above and we have decomposed the off-diagonal correlation into a magnitude $\Delta^{l}$ and phase $\varphi^{l}$. 

Note that the symmetry group of \eqref{2repac} is $U(1)_{a} \times U(1)_{b}$. Formally speaking, provided $\Delta \neq 0$, the {\it mixed} symmetry is spontaneously broken and $\phi$ can be thought of as its Goldstone mode, as we have the following transformation that leaves the action invariant: 
\be
z^{l}_{i,a} \to e^{i\al_{a}} z^{l}_{i,a} \qquad z^{l}_{i,b} \to e^{i\al_{b}} z^{l}_{i,b}
\qquad \varphi^{l} \to -\al_{a} + \al_{b} \label{phi_is_a_goldstone} 
\ee
This nonlinear shift is precisely what defines a Goldstone mode. On general grounds one might then expect the time-dependence of this mode to be slow. We see an extreme version of this: $\phi^{l}$ is actually {\it time-independent}, i.e. 
\be
\varphi^{l+1} = \varphi^{l}
\ee
We will verify this explicitly in a calculation, but it can immediately be seen by realizing that a shift in $\phi^{l=0}$ may be obtained by performing a $U(1)$ rotation of $z_{a}$ as an input. This $U(1)$ rotation must then propagate through each layer by equivariance, rotating the off-diagonal component $C^{l}_{ab}$ in precisely the same way at each layer. 

We now derive an equation for the evolution of $C^{l}_{ab}$. Explicitly, the inverse is:
\be
[C^{-1}]_{ab} = \frac{1}{c^2 - \Delta^2}\begin{pmatrix} 
            c & -\Delta e^{i\varphi} \\
            -\Delta e^{-i\varphi} & c 
            \end{pmatrix}
\ee
The generalization of \eqref{ceqn_zz} now becomes:
\be
C_{ab}^{l+1} = \frac{1}{\det C^{l} (\pi \sig_w^2)^2} \int dZ dZ^{\dagger} \exp\le(-\frac{1}{\sig_w^2} Z^{\dagger} (C^{l})^{-1} Z\ri) \phi^{\dagger}(z_a)\phi(z_b) \label{cab_evolve} 
\ee
where we have combined the two replicas into one vector $Z = (z_1, z_2)$. Change variables as $z_a = r_a e^{i\th_a}$ and then we have
\be
Z^{\dagger} C^{-1} Z = \frac{1}{c^2 - \Delta^2}\le(c^2(r_1^2 + r_2^2) - 2 \Delta r_1 r_2 \cos(\varphi + \th_1 - \th_2)\ri) \label{exponent} 
\ee
We now evaluate \eqref{cab_evolve} at the interesting off-diagonal point $a = 1, b = 2$. 

We first track the evolution of the phase $\phi^{l}$. On the right-hand side it appears only in the exponent through \eqref{exponent}; we can remove it from here by shifting $\th_{1} \to \th_{1} - \varphi^{l}$. The integration measure is invariant under this change of variables and the explicit appearance of $\phi^{\dagger}(z_1)$ transforms equivariantly; the right-hand side thus contains an explicit factor of $e^{i\varphi^{l}}$.  

We now perform the angular integrals. Define the angles $\delta = \th_1 - \th_2$ and $\Sigma = \ha(\th_1 + \th_2)$. The integral over $\Sig$ factors out, and we find an integral over $\delta$ which can be done using Bessel functions. Ultimately only the radial integrals remain, and we find:
\be
\fontsize{7.5pt}{13pt}\selectfont
\Delta^{l+1} e^{i\varphi^{l+1}}= \frac{4 e^{i\varphi^l}}{(c^2-\Delta^2)\sig_{w}^4} \int dr_1 dr_2 r_1 r_2 \exp\le(-\frac{c}{\sig_w^2(c^2 - \Delta^2)}(r_1^2 + r_2^2)\ri) I_1\le(2r_1 r_2 \frac{\Delta}{\sig_{w}^2(c^2 - \Delta^2)}\ri) \tanh(r_1) \tanh(r_2)
\ee
Note, as claimed, that we have that $\varphi^{l+1} = \varphi^{l}$: the Goldstone mode does not evolve across the layers. The remainder of the integral is an evolution equation for the real degree of freedom $\Delta^{l}$.  To make this integral easier to evaluate we perform the change of variables
\be
u_{a} = \frac{\sqrt{c}}{\sig_{w}\sqrt{c^2-\Delta^2}}r_{a} \equiv \al r_{a}
\ee
after which we have:
\be
\Delta^{l+1} = \frac{4(c^2 - \Delta^2)}{c^2} \int du_1 du_2 u_1 u_2 \exp\le(-(u_1^2 + u_2^2)\ri) I_1\le(2u_1 u_2 \frac{\Delta}{c}\ri) \tanh\le(\frac{u_1}{\al}\ri) \tanh\le(\frac{u_2}{\al}\ri) \label{evoleq} 
\ee
where everything on the right-hand side is evaluated at layer $l$.

\subsection{Jacobian of input-output map} \label{app:jacobian}

Here we demonstrate that the following order-parameter
\be
d^l \equiv \frac{1}{N} \sum_{i} \mathbb{E}(z^l_i z^{l\dagger}_{i})
\ee
i.e. essentially the same information as $c^{l}$ in \eqref{cdef}, but evaluated on pre-activations instead -- determines a certain component of the Jacobian of the input-output map. Consider a $U(1)$-equivariant network with $L$ layers. We will study the Jacobian of the output $z^{L}_i$ with respect to the input $z^{0}_i$, i.e. the components of the matrices:
\be
\frac{\p z^L_{i}}{ \p z^0_{j}} \qquad \frac{\p z^L_{i}}{\p z^{0\dagger}_{i}} 
\ee
and their complex conjugates.  

\begin{figure}
    \centering
    \includegraphics[width=0.6\linewidth]{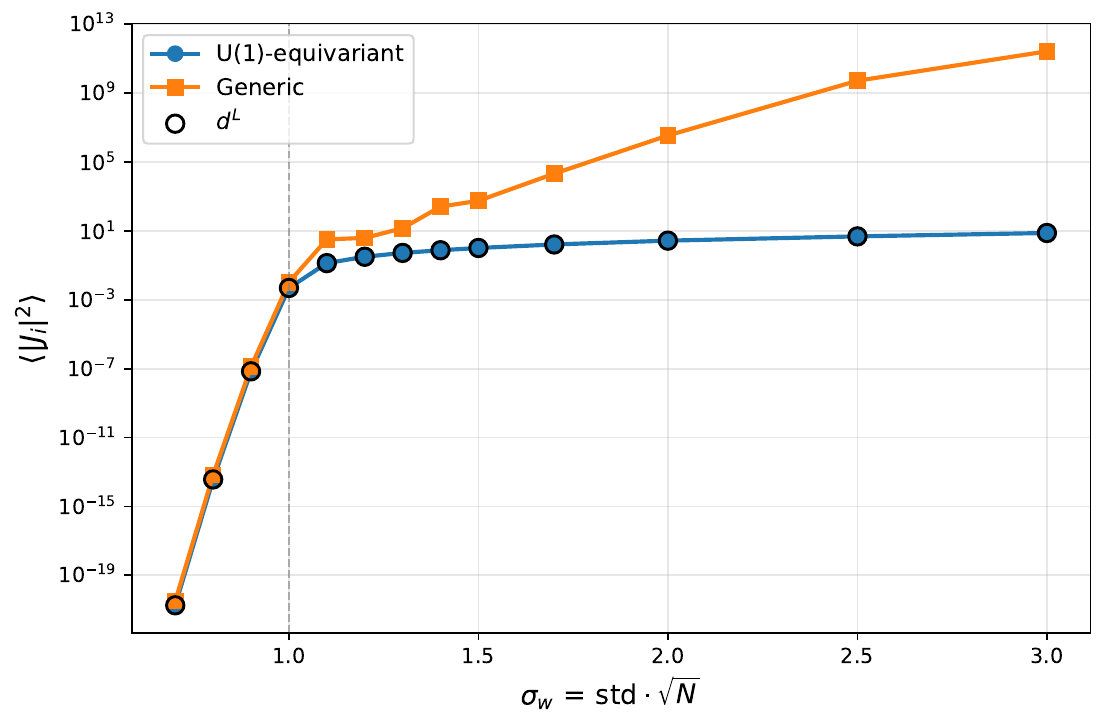}
    \caption{Protected component of Jacobian at initialization. In the symmetry unbroken phase, all components of the Jacobian decay exponentially with layer. In the symmetry broken-phase, one of the components of the equivariant Jacobian can be related to $d^{L}$ (plotted as circles above) and thus remains $O(1)$ for all $\sig_{W}$, whereas the corresponding quantity for the non-equivariant model grows exponentially with $L$. This can be interpreted as the existence of a protected channel for information propagation.} 
    \label{fig:jacobian}
\end{figure}

Due to the equivariance, a $U(1)$ rotation on the input $\delta z^0_{i} = i \ep z^{0}_i$ results in the corresponding transformation of the {\it output} $\delta z^{L}_i = i \ep z^{L}_i$, leading to the following constraint:
\be
i \ep z^{L}_i = \sum_{j} \le(\frac{\p z^{L}_i}{\p z^{0}_j} i \ep z^{0}_j - \frac{\p z^{L}_i}{\p z^{0\dagger}_j} i \ep z^{0\dagger}_j\ri) \equiv i \ep J_{i} \label{Jdef} 
\ee
where we have defined a particular component of the Jacobian -- correlated in a specific fashion with the input $z^{0}_i$ -- to be
\be
J_{i} \equiv \sum_{j} \le(\frac{\p z^{L}_i}{\p z^{0}_j}  z^{0}_j - \frac{\p z^{L}_i}{\p z^{0\dagger}_j}  z^{0\dagger}_j\ri)
\ee
Note that this is an algebraic constraint relating derivatives to the output of the network: one does not expect such a relation unless equivariance is involved. We now multiply both sides of \eqref{Jdef} with their complex conjugate and take the expectation value to find
\be
d^{L} = \frac{1}{N} \sum_{i} \mathbb{E} (|J_i|^2)
\ee
This is a non-trivial relation. To understand why, consider dynamics at initialization. For a conventional neural network in the chaotic (ordered) phase, one expects the Jacobian to grow (decay) exponentially with layer $L$. However, as discussed in the main text, in the presence of equivariance the chaotic phase can also be interpreted as a phase with symmetry breaking, and in that case we see that a {\it particular component} of the Jacobian is controlled by the order parameter $d^{L}$, which is $O(1)$ at large $L$. Thus we see that everywhere in the SSB phase the combination of derivatives constituting $J_{i}$ is $O(1)$, as we verify in Figure \ref{fig:jacobian}. The implications of this for training are discussed around Figure \ref{fig:rank_combined} in the main text. 

We note that our argument -- showing how a symmetry transformation must propagate across distance in the SSB phase, resulting in a long-range correlation -- is very similar to that used to derive the physics version of Goldstone's theorem from the path integral formalism in \cite{Hofman:2018lfz}.

\section{Experimental details}
\label{app:further_experiments}
In the following two subsections, we provide some further experimental details for the MLP investigations that are used to support the theory of Section \ref{sec:model_theory}, and the RNN models and tasks presented in Section \ref{sec:experiments}.

\subsection{MLP experimental implementation details} 
We implement the models in PyTorch \cite{paszke2019pytorch} and used a learning rate of $10^{-3}$ with Adam \cite{kingma2017adammethodstochasticoptimization} for optimization with a batch size of 256. As the idea is to demonstrate the effects of equivariance rather than to optimize over performance, we did not sweep over hyperparameters except to add and remove equivariant architectures, as documented in the text. 

The experiments were very lightweight on toy datasets MNIST and Fashion-MNIST \cite{xiao2017fashion} with the standard splits, and could generally be accomplished on CPU or < 1 hour on a RTX 3090. 

\begin{figure}
    \centering
    \includegraphics[width=0.6\linewidth]{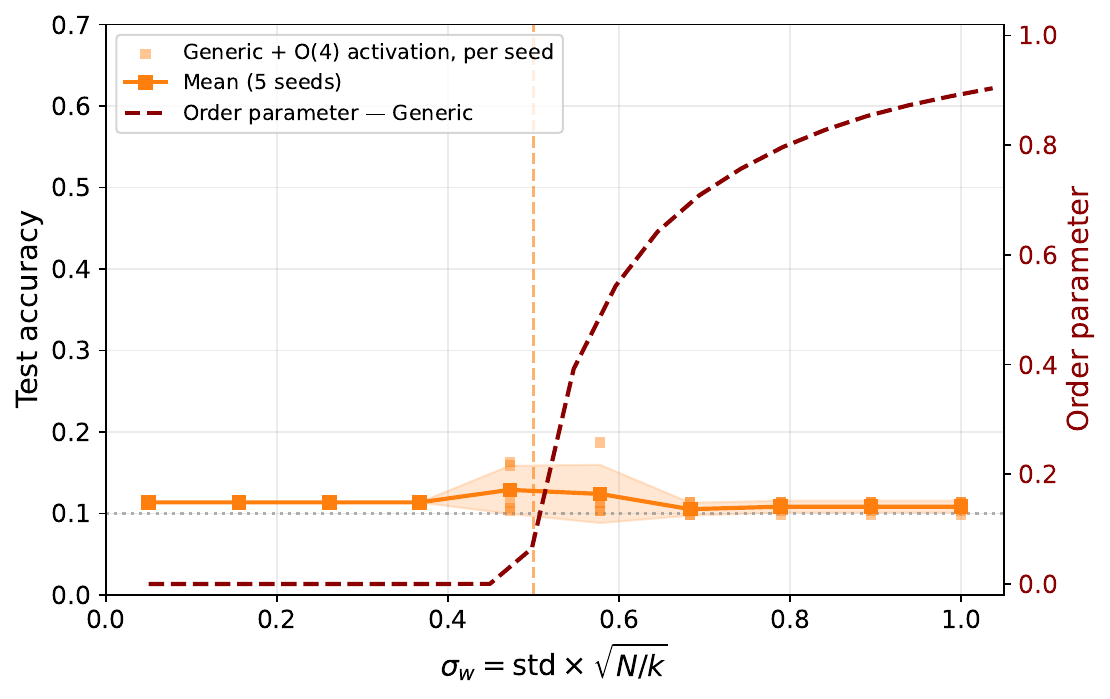}
    \caption{Here we demonstrate the ``conventional'' edge of chaos for a model with the same nonlinearity as in the right panel of Figure \ref{fig:test-accuracy-phasetransition} but with generic linear layers that do not preserve the $O(4)$ activation. This is zoomed in on the region of the transition, which is at  the left edge of Figure \ref{fig:test-accuracy-phasetransition}. Note that the order parameter is still activated above the transition, but it is {\it not} accompanied by any performance gain, which only happens modestly near the transition itself, in agreement with classical signal propagation theory. This highlights that it is the equivariance (which allows nontrivial modulations of the order parameter by the symmetry group) that is crucial for the training advantage in the SSB phase.} 
    \label{fig:generic_eoc}
\end{figure}

\begin{figure}
    \centering
    \begin{subfigure}[t]{0.49\linewidth}
        \centering
        \includegraphics[width=\linewidth]{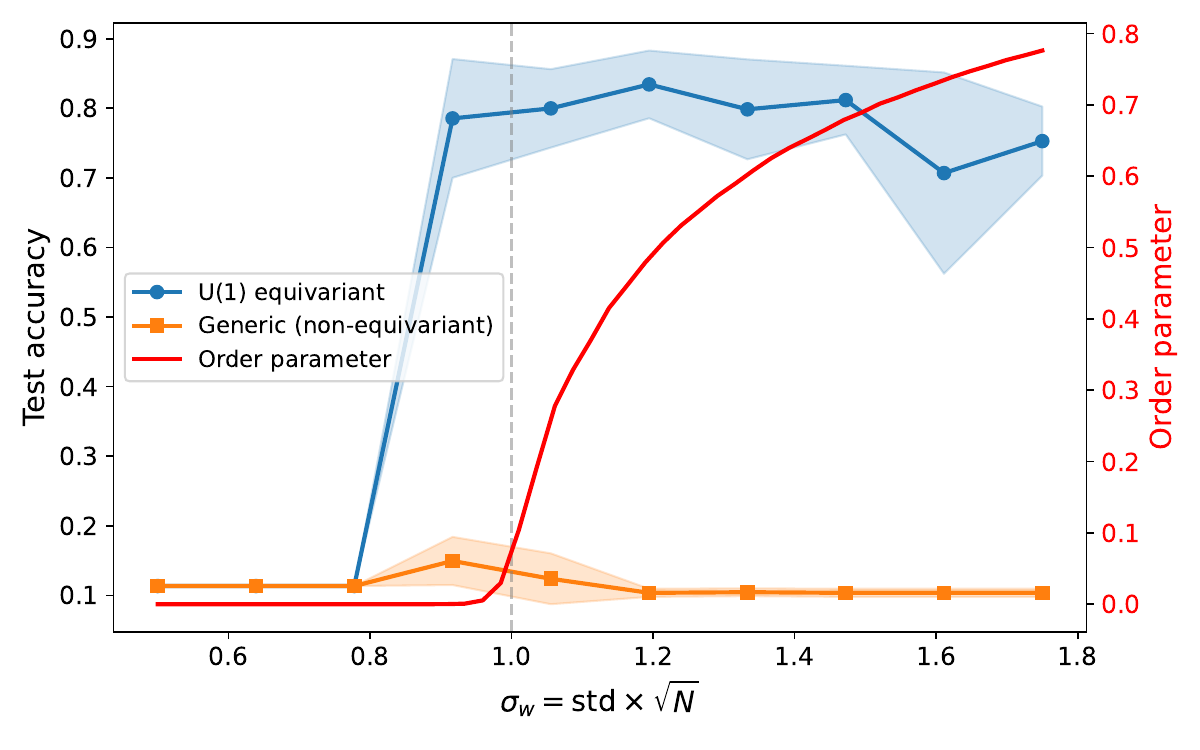}
        %\caption{$U(1)$ equivariant fully connected model with 64 features and 100 layers. Note that the model with no equivariance does not train at all, except slightly in the vicinity of the phase transition (the usual ``edge of chaos'').}
        \label{fig:test-accuracy-phasetransition-u1-app}
    \end{subfigure}
    \hfill
    \begin{subfigure}[t]{0.49\linewidth}
        \centering
        \includegraphics[width=\linewidth]{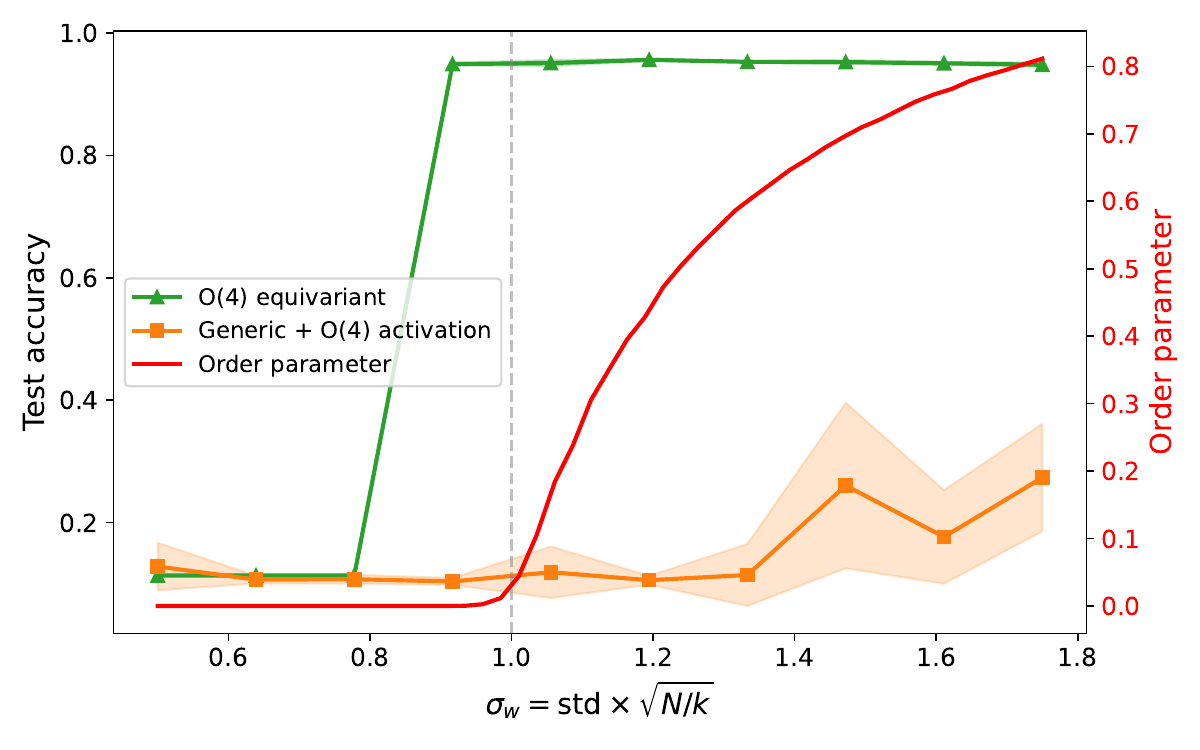}
        %\caption{$O(4)$ equivariant fully connected model with 64 features and 100 layers.}
        \label{fig:test-accuracy-phasetransition-o4-app}
    \end{subfigure}
    \caption{Test accuracy after 5 epochs on MNIST (as compared to Fashion-MNIST in Figure \ref{fig:test-accuracy-phasetransition}) for $U(1)$ (left) and $O(4)$ (right) equivariant  models. We compare against models with the same nonlinearity but generic linear layers, thus isolating the effect of the equivariance. We vary $\sig_{w}$ to probe the symmetry breaking phase transition. The training of the equivariant models appears to begin precisely when we enter the SSB phase, as measured by the order parameter $c^{\star}$, which we compute at {\it initialization} from \eqref{cevolve}. Training of the non-equivariant models is poor. These results are essentially the same as in Figure \ref{fig:test-accuracy-phasetransition} except that the numbers are slightly higher as the task is easier. }
    \label{fig:test-accuracy-phasetransition-app}
\end{figure}

In Figures \ref{fig:test-accuracy-phasetransition}, \ref{fig:test-accuracy-phasetransition-app}, and  \ref{fig:generic_eoc} we used $N = 64$ features and trained for 5 epochs with 5 seeds. The order parameter is computed at initialization and averaged over 100 instantiations. 

In Figure \ref{fig:fmnist_with_layers} we used $Nk = 128$ and initialized with a fixed variance for the initial weights of $0.20$, which is always deep in the SSB phase for all architectures. We trained for 4 epochs and show results averaged over 3 seeds. 

In Figure \ref{fig:rank_combined} we plot an effective rank by computing $\exp(H)$, where $H$ is an entropy computed from the singular values $\sig_i$ of the representations in (batch, feature) space as in \cite{daneshmand2020batch} 
\be                                                                                                                                            
  p_i \equiv \frac{\sigma_i}{\sum_j \sigma_j} \qquad                                                                                                                                          
  H \equiv - \sum_{i:,\sigma_i > 0} p_i ,\log p_i,
\ee                                                      

In the left panel of Figure \ref{fig:rank_combined} we use $N = 64$ and $L = 64$, fixing $\sig_{w} = 1.5$ to be deep in the SSB phase and training over one epoch. In the right panel, which is over a single seed, we set $N = 128$ and $L = 90$ to obtain a wider range of layers.

\subsection{RNN experimental implementation details} 
% \NI{should we maybe merge this into above}
\label{app:experiment_details}
Below we provide the full details for the copy task and sequential image classification experiments.

% \subsubsection{Models}

% The simple RNN model is formally defined as...

% The U(1) Simple RNN... 

% The O(k) RNN ...

% The GRU model is given as ...

% The U(1) GRU ...

\subsubsection{Models}
\label{app:rnn_model_details}
We use $N$ to denote the number of (potentially) symmetry-carrying hidden units. $U(1)$ models have $2N$ real hidden coordinates and $O(k)$ models have $Nk$. The inputs come as a sequence of real vectors $\mathbf{x}^t\in\mathbb{R}^{d_{\rm in}}$ for $t \in [0, L]$, and all recurrent models are initialized with $\mathbf{h}^0=\mathbf{0}$. For the copy task, a linear readout is applied to the hidden state at every timestep to predict the target sequence; for permuted sequential MNIST, the readout is applied only to the final hidden state to predict the digit class.

\paragraph{Baseline RNNs.}
The non-equivariant simple RNN baseline is a standard one-layer Elman RNN,
\begin{equation}
    \mathbf{h}^t
    =
    \psi\!\left(
        \mathbf{W}_x\mathbf{x}^t
        +
        \mathbf{W}_h\mathbf{h}^{t-1}
        +
        \mathbf{b}
    \right),
    \qquad
    \hat{\mathbf{y}}^t = \mathbf{R}\mathbf{h}^t+\mathbf{c},
    \label{eq:app_baseline_rnn}
\end{equation}
where $\psi\in\{\tanh,\mathrm{ReLU}\}$ is selected by validation and $\mathbf{R}$ is the linear readout. For both tasks, the output $\hat{\mathbf{y}}^t$ is in $\mathbb{R}^{10}$, treated as a vector of logits, and passed through a softmax to predict the current target one-hot vector.

\paragraph{Baseline GRUs.} The GRU baseline is a standard one-layer GRU,
\begin{equation}
\begin{aligned}
    \mathbf{r}^t &=
    \sigma\!\left(
        \mathbf{W}_{xr}\mathbf{x}^t
        +
        \mathbf{W}_{hr}\mathbf{h}^{t-1}
        +
        \mathbf{b}_r
    \right),\\
    \mathbf{z}^t &=
    \sigma\!\left(
        \mathbf{W}_{xz}\mathbf{x}^t
        +
        \mathbf{W}_{hz}\mathbf{h}^{t-1}
        +
        \mathbf{b}_z
    \right),\\
    \tilde{\mathbf{h}}^t &=
    \mathrm{ReLU}\!\left(
        \mathbf{W}_{xn}\mathbf{x}^t
        +
        \mathbf{b}_{xn}
        +
        \mathbf{r}^t\odot
        \left(
            \mathbf{W}_{hn}\mathbf{h}^{t-1}
            +
            \mathbf{b}_{hn}
        \right)
    \right),\\
    \mathbf{h}^t &=
    \mathbf{z}^t\odot\mathbf{h}^{t-1}
    +
    (1-\mathbf{z}^t)\odot\tilde{\mathbf{h}}^t .
\end{aligned}
\label{eq:app_baseline_gru}
\end{equation}
Where the baseline readout is again a learned real affine map from the hidden state to the task output with the same non-linearities as for the copy task.

\paragraph{Equivariant input embeddings.}
For the equivariant models, the real input is first embedded into the hidden representation on which the symmetry acts. For $U(1)$ models, this is implemented as a learned complex embedding
\begin{equation}
    \mathbf{u}^t
    =
    \mathbf{E}_{\mathrm{Re}}\mathbf{x}^t
    +
    i\,\mathbf{E}_{\mathrm{Im}}\mathbf{x}^t
    \in \mathbb{C}^{N},
\end{equation}
where $\mathbf{E}_{\mathrm{Re}}, \mathbf{E}_{\mathrm{Im}} \in \mathbb{R}^{N \times d_{in}}$. For $O(k)$ models, the embedding matrix $\mathbf{E}_{\mathrm{O(k)}} \in \mathbb{R}^{N k \times d_{in}}$ has $k$ times greater output dimensions, and the output is reshaped into $N$ real $k$-vectors,
\begin{equation}
    \mathbf{u}^t_{\alpha}\in\mathbb{R}^k,
    \qquad
    \alpha=1,\ldots,N.
\end{equation}
% The input and output maps are task-specific real-valued maps and the equivariance is imposed on the hidden recurrent computation.

\paragraph{$U(1)$-equivariant RNN.}
The $U(1)$ hidden state is complex, $\mathbf{h}^t\in\mathbb{C}^N$, designed to be equivariant to the global action $\mathbf{h}^t\mapsto e^{i\theta}\mathbf{h}^t$. The recurrent map is complex-linear and the
nonlinearity is radial:
\begin{equation}
    \mathbf{h}^t
    =
    \phi_{U(1)}
    \!\left(
        \mathbf{u}^t
        +
        \mathbf{W}\mathbf{h}^{t-1}
    \right),
    \qquad
    \mathbf{W}\in\mathbb{C}^{N\times N},
    \label{eq:app_u1_rnn}
\end{equation}
with
\begin{equation}
    \left[\phi_{U(1)}(\mathbf{z})\right]_\alpha
    =
    \frac{\tanh(|z_\alpha|)}{|z_\alpha|+\epsilon}
    z_\alpha.
    \label{eq:app_u1_radial}
\end{equation}
Thus the activation changes the magnitude of each complex hidden coordinate while preserving its phase. In practice we use $\epsilon = 10^{-8}$. The reported $U(1)$ RNN and GRU runs use a real readout
\begin{equation}
    \hat{\mathbf{y}}^t
    =
    \mathbf{R}
    \begin{bmatrix}
        \operatorname{Re}[\mathbf{h}^t]\\
        \operatorname{Im}[\mathbf{h}^t]
    \end{bmatrix}
    +
    \mathbf{c},
    \label{eq:app_u1_readout}
\end{equation}
which maps the complex hidden representation back to the real task output space.

\paragraph{$O(k)$-equivariant RNN.}
For $O(k)$ models, the hidden state is $\mathbf{h}^t\in\mathbb{R}^{N\times k}$, where each $\mathbf{h}^t_\alpha\in\mathbb{R}^k$
is designed to be equivariant to the action 
\begin{equation}
    \mathbf{h}^t_\alpha \mapsto O\mathbf{h}^t_\alpha,
    \qquad
    O\in O(k),
\end{equation}
with the same orthogonal matrix acting on every hidden block indexed by $\alpha$. The recurrent update is
\begin{equation}
    \mathbf{q}^t_{\alpha a}
    =
    \mathbf{u}^t_{\alpha a}
    +
    \sum_{\beta=1}^{N}
    \mathbf{A}_{\alpha\beta}\mathbf{h}^{t-1}_{\beta a},
    \qquad
    \mathbf{h}^t_{\alpha}
    =
    \phi_{O(k)}(\mathbf{q}^t_\alpha),
    \label{eq:app_ok_rnn}
\end{equation}
where $\mathbf{A}\in\mathbb{R}^{N\times N}$. In flattened coordinates the recurrent matrix is $\mathbf{A}\otimes I_k$, so it mixes the $N$ hidden blocks but acts identically on each of the $k$ internal coordinates. The nonlinearity acts only through the $O(k)$-invariant radius:
\begin{equation}
    \phi_{O(k)}(\mathbf{v}_\alpha)
    =
    \frac{\tanh(\|\mathbf{v}_\alpha\|_2)}{\|\mathbf{v}_\alpha\|_2+\epsilon}
    \mathbf{v}_\alpha.
    \label{eq:app_ok_radial}
\end{equation}
Equivalently, this replaces the length of each $k$-vector by the scalar nonlinearity of its length, while preserving its direction in $\mathbb{R}^k$. The implementation allows either an invariant norm readout,
\begin{equation}
    \hat{\mathbf{y}}^t
    =
    \mathbf{R}\mathbf{s}^t+\mathbf{c},
    \qquad
    s^t_\alpha=\|\mathbf{h}^t_\alpha\|_2,
\end{equation}
or a flattened real readout, $\hat{\mathbf{y}}^t=\mathbf{R}\operatorname{vec}(\mathbf{h}^t)+\mathbf{c}$. In practice, we find the flattened readout to perform best and thus use it in all reported experiments.

\paragraph{$U(1)$-equivariant GRU.}
The $U(1)$ GRU uses the same complex hidden state and radial candidate update as the $U(1)$ RNN, but adds real-valued reset and update gates. The gates are computed from the real input and from
phase-invariant hidden features:
\begin{equation}
\begin{aligned}
    \mathbf{r}^t
    &=
    \sigma\!\left(
        \mathbf{W}_{xr}\mathbf{x}^t
        +
        \mathbf{W}_{hr}|\mathbf{h}^{t-1}|
        +
        \mathbf{b}_r
    \right),\\
    \mathbf{z}^t
    &=
    \sigma\!\left(
        \mathbf{W}_{xz}\mathbf{x}^t
        +
        \mathbf{W}_{hz}|\mathbf{h}^{t-1}|
        +
        \mathbf{b}_z
    \right),\\
    \tilde{\mathbf{h}}^t
    &=
    \phi_{U(1)}
    \!\left(
        \mathbf{u}^t
        +
        \mathbf{W}_{h}
        \left(
            \mathbf{r}^t\odot\mathbf{h}^{t-1}
        \right)
    \right),\\
    \mathbf{h}^t
    &=
    \mathbf{z}^t\odot\mathbf{h}^{t-1}
    +
    (1-\mathbf{z}^t)\odot\tilde{\mathbf{h}}^t .
\end{aligned}
\label{eq:app_u1_gru}
\end{equation}
Here $|\mathbf{h}|$ is the complex magnitude, taken elementwise, $\sigma$ denotes a sigmoid nonlinearity, $\odot$ denotes the elementwise product, and $\mathbf{r}^t,\mathbf{z}^t\in[0,1]^N$ are real. Since the gates depend on the hidden state only through phase-invariant magnitudes, multiplying $\mathbf{h}^{t-1}$ by a global phase leaves the gates unchanged, while the candidate branch transforms equivariantly. We use the default gate initialization $\mathbf{b}_r=\mathbf{0}$ and $\mathbf{b}_z=2$, so the initialized update gate is biased toward copying the previous hidden state.

\paragraph{Parameter counts.}
All parameter counts in Tables \ref{tab:optimal_params}, \ref{tab:results_table_copy} \& \ref{tab:results_table_psmnist} and Figures \ref{fig:copy_joint} \& \ref{fig:psmnist} report trainable real
scalar parameters. Thus each trainable complex scalar counts as two real
parameters. 
% Let $d_{\rm in}$ and $d_{\rm out}$ denote the input and output
% dimensions, with $(d_{\rm in},d_{\rm out})=(10,10)$ for copy and $(1,10)$ for
% psMNIST.
% With the implementation biases used in our recurrent baselines, the
% parameter counts are
% % 
% \begin{equation}
%     \begin{aligned}
% P_{\rm RNN}(N)
% &= N^2 + N d_{\rm in} + 2N + d_{\rm out}N + d_{\rm out},\\
% P_{\rm GRU}(N)
% &= 3N^2 + 3N d_{\rm in} + 6N + d_{\rm out}N + d_{\rm out},\\
% P_{U(1)\text{-RNN}}(N)
% &= 2N^2 + 2N d_{\rm in} + 2N d_{\rm out} + d_{\rm out},\\
% P_{U(1)\text{-GRU}}(N)
% &= 4N^2 + 4N d_{\rm in} + 2N d_{\rm out} + 2N + d_{\rm out}.
% \end{aligned}
% \end{equation}
% % 
% For the $U(1)$ GRU, the terms are respectively the complex input embedding,
% two real gates, the complex candidate recurrent map, gate biases, and the real
% readout:
% \begin{equation}
% 2Nd_{\rm in} + (2Nd_{\rm in}+2N^2+2N) + 2N^2
% + (2Nd_{\rm out}+d_{\rm out}).
% \end{equation}

\subsubsection{Initialization and hyperparameter sweeps.}
For all models we initialize the hidden state to zero at the start of each sequence: $\mathbf{h}^0=\mathbf{0}$. We sweep over identity and random recurrent initializations. Identity initialization corresponds to $\mathbf{W} =I_N \in \mathbb{C}^N$ for the $U(1)$ RNN, $\mathbf{A}=I_N$ for the $O(k)$ RNN, and identity recurrent blocks for the baseline recurrent models. Random recurrent initialization uses the default PyTorch initialization corresponding to sampling weights uniformly at random from $U(-\frac{1}{\sqrt{N}}, \frac{1}{\sqrt{N}})$. For the GRUs, identity initialization corresponds to setting $\mathbf{W}_{hr}, \mathbf{W}_{hz}$ and $\mathbf{W}_{h}$ to identity matrices. We additionally sweep over different random initializations for the encoder matrix $\mathbf{E}$ for all models. We consider the standard initialization, where again all weights are sampled uniformly at random from $U(-\frac{1}{\sqrt{N}}, \frac{1}{\sqrt{N}})$, and a second low-variance initialization where the weights are sampled from $\mathcal{N}(0, 0.001)$. For the copy task, with the low-variance initialization, we additionally fix the read-in weight for the `blank' token presented during the delay period to be a vector of all zeros throughout training. For the equivariant models, we find that this is beneficial to preserve the information channel provided by spontaneous symmetry breaking, while we find it has little to no effect on the baseline models (but still include it for fair comparison). For the $O(k)$ RNN on psMNIST, we performed a random search over values of neurons $N \in \{64, 128, 256, 512\}$, and $k \in \{6, 8, 12, 18, 24, 48\}$ up to our compute budget, and report the best performing combination at each parameter range. For the reported results we additionally sweep over learning rates from the set $\{10^{-3}, 10^{-4}, 10^{-5}\}$. 

\looseness=-1
\paragraph{Best performing hyperparameters.} 
From the sweeps reported above, we report the hyperparameters found to be optimal for each architecture, dataset, and hidden state size in Table \ref{tab:optimal_params} below.

\begin{table}[h]
\centering
% \small
% \setlength{\tabcolsep}{4pt}
% \renewcommand{\arraystretch}{1.08}
\caption{
Best-performing hyperparameters selected by validation performance for each dataset, architecture, and hidden size. We sweep recurrent initialization, input/encoder initialization, activation function, and learning rate (and $k$ when appropriate), and report the setting with the best mean validation performance across sweep seeds.
}
\vspace{3mm}
\label{tab:optimal_params}
\begin{tabular}{llclllll|c}
\toprule
Dataset & Model      & $N$ & Rec. init. & Input init. & Activation & LR & $k$ & \# Params \\
\midrule
Copy    & RNN        & 32  & identity & standard  & $\tanh$ & $10^{-4}$ & -  & 1.7K \\
Copy    & RNN        & 64 & identity & standard  & $\tanh$ & $10^{-4}$ &  - & 5.5K \\
Copy    & GRU        & 32  & identity   & standard  & ReLU    & $10^{-3}$ & -  & 4.5K \\
Copy    & GRU        & 64 & identity   & low-var.  & ReLU    & $10^{-3}$ &  - & 15K \\
Copy    & $U(1)$ RNN & 32  & identity & low-var.  & radial $\tanh$ & $10^{-3}$ & - & 3.3K \\
Copy    & $U(1)$ GRU & 32  & random & low-var.   & radial $\tanh$ & $10^{-3}$ & -  & 6.1K \\
\midrule
psMNIST & RNN        & 128  & identity   & standard  & ReLU    & $10^{-4}$ &  - & 18K \\
psMNIST & RNN        & 256  & identity   & standard  & ReLU    & $10^{-5}$ &  - &  69K \\
psMNIST & RNN        & 512  & identity   & standard  & ReLU    & $10^{-4}$ &  - & 269K \\
psMNIST & RNN        & 1024 & identity   & standard  & ReLU    & $10^{-4}$ &  - & 1.06M \\
psMNIST & GRU        & 64   & identity   & standard  & ReLU    & $10^{-3}$ & -  & 14K \\
psMNIST & GRU        & 128  & identity   & standard  & ReLU    & $10^{-3}$ & -  & 52K \\
psMNIST & GRU        & 256  & identity   & standard  & ReLU    & $10^{-3}$ & -  & 200K \\
psMNIST & GRU        & 512  & identity   & standard  & ReLU    & $10^{-3}$ &  - & 796K \\
psMNIST & $O(k)$ RNN & 64  & identity & low-var. & radial $\tanh$ & $10^{-4}$ & 18 & 18k \\
psMNIST & $O(k)$ RNN & 128 & identity & low-var. & radial $\tanh$ & $10^{-4}$ & 8 & 29k \\
psMNIST & $O(k)$ RNN & 256 & identity & low-var. & radial $\tanh$ & $10^{-4}$ & 6 & 84k \\
psMNIST & $O(k)$ RNN & 256 & identity & low-var. & radial $\tanh$ & $10^{-4}$ & 24 & 139k \\
psMNIST & $O(k)$ RNN & 256 & identity & low-var. & radial $\tanh$ & $10^{-4}$ & 48 & 213k \\
psMNIST & $O(k)$ RNN & 512 & identity & low-var. & radial $\tanh$ & $10^{-4}$ & 24 & 410k \\
psMNIST & $U(1)$ GRU & 64 & identity & low-var. & radial $\tanh$ & $10^{-3}$ &  - & 18K \\
psMNIST & $U(1)$ GRU & 128 & identity & low-var.  & radial $\tanh$ & $10^{-3}$ & - & 69K \\
psMNIST & $U(1)$ GRU & 256  & identity & low-var. & radial $\tanh$ & $10^{-3}$ & - & 269K \\
psMNIST & $U(1)$ GRU & 512 & identity & low-var.  & radial $\tanh$ & $10^{-3}$ & -  & 1.06M \\
\bottomrule
\end{tabular}
\end{table}

\looseness=-1
\subsubsection{Extended Results.} 
In Tables \ref{tab:results_table_copy} and \ref{tab:results_table_psmnist}, we provide the exact values for the data presented in Figures \ref{fig:copy_joint} and \ref{fig:psmnist}. In Figure \ref{fig:copy_joint}, the significance stars are calculated via a pairwise Welch’s two-sample t-test between each pair of models.  

\begin{table}[h!]
\centering
% \small
% \setlength{\tabcolsep}{4pt}
% \renewcommand{\arraystretch}{1.08}
% \caption{
% Summary of the results presented in Figure \ref{fig:copy_joint}. Mean $\pm$ standard deviation over 10 random seeds.
% }
\caption{Summary of the variable-delay copy task results presented in Figure \ref{fig:copy_joint}. We report test cross-entropy loss, averaged over sequence length and examples. Values are mean $\pm$ standard deviation over 10 random seeds.}
\vspace{3mm}
\label{tab:results_table_copy}
\begin{tabular}{llc|cc}
\toprule
$T_{max}$ & Model      & $N$ &  Test CE loss $\pm$ Stdev. & \# Params \\
\midrule
25    & RNN        & 32  & 8.40e-3 $\pm$ 4.39e-4  & 1.7K \\
25    & RNN        & 64  & 2.08e-3 $\pm$ 3.84e-4  & 5.5K \\
25    & $U(1)$ RNN & 32  & \textbf{7.26e-4} $\pm$ \textbf{5.47e-4}  & 3.3K \\
\midrule
100    & GRU        & 32  & 2.11e-3 $\pm$ 4.02e-4  & 4.5K \\
100    & GRU        & 64  & 3.17e-3 $\pm$ 4.03e-3  & 15K \\
100    & $U(1)$ GRU & 32  &  \textbf{3.14e-4} $\pm$ \textbf{4.41e-4} & 6.1K \\
\bottomrule
\end{tabular}
\end{table}

\begin{table}[h]
\centering
% \small
% \setlength{\tabcolsep}{4pt}
% \renewcommand{\arraystretch}{1.08}
\caption{
Summary of the psMNIST results presented in Figure \ref{fig:psmnist}. We report test set accuracy. Values are mean $\pm$ standard deviation over 3 random seeds.
}
\vspace{3mm}
\label{tab:results_table_psmnist}
\begin{tabular}{lcc|cc}
\toprule
 Model      & $N$ & $k$ & Test Acc. $\pm$ Stdev. & \# Params \\
\midrule
RNN        & 128  & - & 85.19 $\pm$ 3.13 & 18K \\
RNN        & 256  & - & 88.90 $\pm$ 0.73 & 69K \\
RNN        & 512  & - & 90.92 $\pm$ 1.02 & 269K \\
RNN        & 1024 & - & 91.00 $\pm$ 2.57 & 1.06M \\
$O(k)$ RNN & 64   & 18 & 87.09 $\pm$ 0.55 & 18K \\
$O(k)$ RNN & 128  & 8  & 88.98 $\pm$ 0.92 & 29K \\
$O(k)$ RNN & 256  & 6  & 92.01 $\pm$ 1.43 & 84K \\
$O(k)$ RNN & 256  & 24 & 93.34 $\pm$ 1.86 & 139K \\
$O(k)$ RNN & 256  & 48 & 94.16 $\pm$ 0.23 & 213K \\
$O(k)$ RNN & 512  & 24 & 94.79 $\pm$ 0.22 & 410K \\
\midrule   
GRU        & 64   & - & 91.35 $\pm$ 1.88 & 14K \\
GRU        & 128  & - & 92.77 $\pm$ 0.99 & 52K \\
GRU        & 256  & - & 94.41 $\pm$ 0.63 & 201K \\
GRU        & 512  & - & 95.08 $\pm$ 0.35 & 796K \\
$U(1)$ GRU & 64   & - & 93.51 $\pm$ 0.86 & 18K \\
$U(1)$ GRU & 128  & - & 94.92 $\pm$ 0.39 & 69K \\
$U(1)$ GRU & 256  & - & 95.88 $\pm$ 0.22 & 269K \\
$U(1)$ GRU & 512  & - & 96.21 $\pm$ 0.14 & 1.06M \\
\bottomrule
\end{tabular}
\end{table}

\subsubsection{Datasets}
\paragraph{Variable-Delay Copy Task.}
We use a variable-delay version of the standard copy task~\cite{hochreiter1997long, graves2014neural,henaff2016recurrent,gu2020improving}. Each example begins with a length-10 sequence of randomly sampled one-hot data tokens which the model must store in memory. The model is then presented with a sequence of blank delay tokens, followed by a special go token which indicates that the stored sequence should be reproduced. Each token is a 10-dimensional one-hot vector, where the delay token is index zero, the memory sequence is randomly sampled from indices 1 to 8, and the `go' token is fixed to index 9. Unlike the fixed-delay version of the task, the delay length is sampled independently for each example as $T \sim \mathcal{U}\{0,\ldots,T_{\max}\}$. The target is set to the blank token at every timestep except for the 10 timesteps following the go token, where the network must output the original sequence in the same order. When batching examples with different delays, sequences are padded with blank tokens up to the maximum length for the corresponding value of $T_{\max}$, and the padded positions are assigned blank targets. Given the limited nature of the synthetic dataset, no fixed training, validation, or test set are created. Instead, fresh batches are randomly sampled for each training, validation, and test iteration. The test batch size is set to 50, and evaluated every 1000 training iterations. 

The variable delay modification prevents the model from relying on a fixed output time, and instead requires it to use the go token to determine when the memory should be read out. We found that without this modification, simple linear RNNs outperformed all non-linear models, making the task less relevant to more real-world problems where non-linearity is almost always required. In the experiments of Figure \ref{fig:copy_joint}, we use $T_{\max}=25$ for the simple RNN experiments and $T_{\max}=100$ for the GRU experiments; we additionally consider $T_{\max}=200$ with the same hyperparameter sweep and find that while nearly all of the models struggle, the $U(1)$ GRU model is the only one to have a single seed which successfully `solves' the task (achieving a test cross-entropy loss of less than $10^{-4}$) and achieves a loss of $3.9 \times 10^{-7}$.

\paragraph{Permuted Sequential MNIST.}
For the sequential image classification experiments, we use the standard permuted sequential MNIST
benchmark~\cite{le2015simple,arjovsky2016unitary}. Each $28 \times 28$ MNIST image is flattened
into a sequence of 784 scalar pixel values and presented to the recurrent model one pixel at a time.
The model receives no additional spatial information, and the digit class is predicted only from the
final hidden state after the full sequence has been processed.

To make the task more difficult than standard sequential MNIST, we sample a single random
permutation of the 784 pixel locations before training and apply this same fixed permutation to every
image in the training, validation, and test sets. This removes the short-range spatial correlations
present in the usual raster ordering of the pixels, forcing the model to integrate information over the
full sequence length rather than exploiting local image structure. We use the standard MNIST
train/test split, with model selection performed on held-out validation performance and final numbers
reported on the test set. The validation set is built from a random 3,000 samples from the original 60,000 sample train set. 

\subsubsection{Training and Evaluation}
We use the Adam optimizer \cite{kingma2017adammethodstochasticoptimization} for all experiments with $\beta_1 = 0.9$ and $\beta_2=0.999$ (PyTorch defaults). 

\paragraph{Copy task.} All copy task models were trained for 1 million iterations with a batch size of $128$. The models are trained and evaluated using cross-entropy loss at each timestep, averaged over the full sequence length and over the batch. Thus, the copy-task losses reported in Figure \ref{fig:copy_joint} and Table \ref{tab:results_table_copy} correspond to test cross-entropy, averaged over sequence length, test batches and random seeds. We run 3 random seeds for each model from the sweep, totaling to 72 runs for each model and hidden state size (2 recurrent initializations, 2 input initializations, 2 activation functions, 3 learning rates, 3 seeds). For the copy task we identified the best performing hyperparameters from this initial sweep (reported in Table \ref{tab:optimal_params}) based on mean validation performance and ran an additional 7 random seeds for the best performing models to get the full distributions presented in Figure \ref{fig:copy_joint}. In full, for the copy task, this resulted in 478 trained models. 

\paragraph{Permuted Sequential MNIST.} All psMNIST models are trained for $120$ epochs with a batch size of $128$. The models are trained with a cross-entropy loss over the 10 MNIST digit class labels, and the loss is averaged over the batch. We use gradient clipping and a learning rate decay schedule found to be optimal from prior work \cite{keller2023traveling} using similar RNNs on this task. Specifically, we used gradient clipping by magnitude with a value of $1.0$, and a step-based learning rate schedule where the learning rate decreases by a factor of $10$ after $100$ epochs. 

\paragraph{Computational requirements.} We ran experiments on a combination of H100 and A100 NVIDIA GPUs. Each model presented here comfortably fits within a single A100 GPU's memory (40GB), with all but the largest $O(k)$ RNNs using less than a few GB.  For the copy task, with $T_{max}=25$, the average runtime for the RNN and U(1) RNNs were similar at around 8 hours for 1M iterations. For the GRU models, the runtime comparison was confounded by a non-optimized implementation of the $U(1)$ GRU causing it to run significantly slower than the non-equivariant baseline ($\approx$ 2 days per $U(1)$ model vs. $\approx$ 4 hours per generic GRU). In total we estimate around 4,000 GPU hours were used to produce the copy task results. For the psMNIST the average runtime was around 2.5 hours for the RNN models, 7 hours for the $O(k)$ RNN models, 14 hours for the $U(1)$ GRU models, and around 1 hour for the generic GRU models. Again, these runtimes are with highly unoptimized implementations in a relatively uncontrolled setting (shared machines) where performance may vary. In theory, we see no reason why an optimized implementation of the equivariant models should not run with the same computational complexity as an equivalent non-equivariant model with the same hidden state dimensionality. In total, we estimate around 4,500 GPU hours were used to produce the psMNIST results. 

\subsection{Additional Results: 2D Convolutional RNN Topological Defects}
\label{app:topo_defects}
In the following we present visualizations of the hidden state dynamics for a 2D convolutional RNN with $U(1)$ symmetry trained on the copy task with a fixed delay length of $T=100$. The model is identical to the $U(1)$-equivariant RNN described in Section \ref{app:experiment_details} above, except the complex-linear hidden to hidden recurrent connectivity matrix is replaced with a complex convolutional kernel of size $3\times3$ with $16$ channels. The spatial dimension of hidden state vector is of size $N=400$, and shaped into a $20\times20$ regular grid, over which the 2D convolution operates. The model has 16 convolutional channels, of which two are plotted below. This model learns to solve the copy task well. We plot the magnitude of the complex hidden state neurons over sequence length in the top row, and we can see that vortices emerge in the hidden state (bottom row), as measured by computing the winding numbers for each point in the phase plot in the middle row. As would be expected, we find that vortices always come in positive and negative pairs (defined by inverse winding directions), and can be observed to annihilate with one another on contact. While we have yet to demonstrate convincingly that these vortices are used to store memory in the model, it is promising that they emerge during the memory storage period of the task, i.e. the first 10 steps, and remain stable over the remaining sequence length. We see in the two plots that different channels learn to have different densities of vortices.

\begin{figure}[h!]
    \centering
    \includegraphics[width=\linewidth]{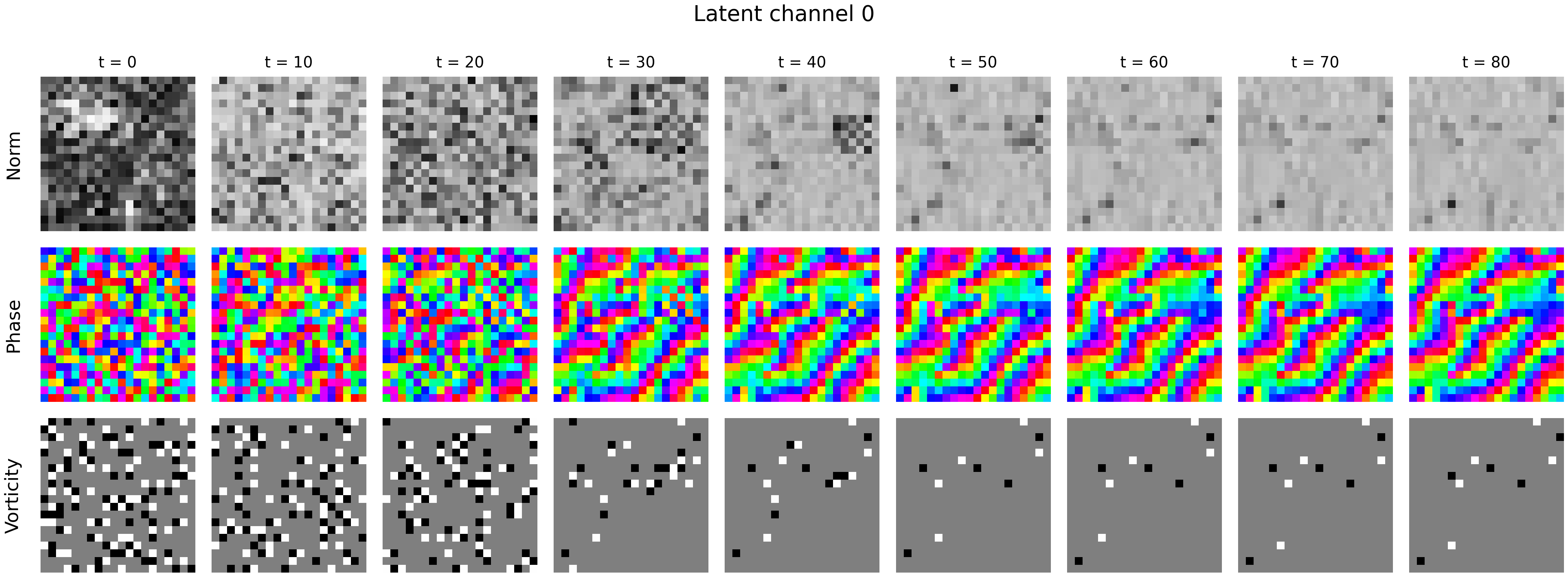}
    \caption{Visualization of hidden state magnitude (top), phase (middle), and vorticity (bottom) for a 2D convolutional U(1) equivariant RNN trained on the copy task with a delay length of 100 and a hidden state size of 400. We show a single channel which appears to exhibit few long lasting vortices which are robustly conserved over the iterations after settling in. Note vortices always come in positive/negative pairs and can annihilate on contact. We find that larger hidden states allow more vortices to persist for longer.}
    \label{fig:vortices_0}
\end{figure}

\begin{figure}[h!]
    \centering
    \includegraphics[width=\linewidth]{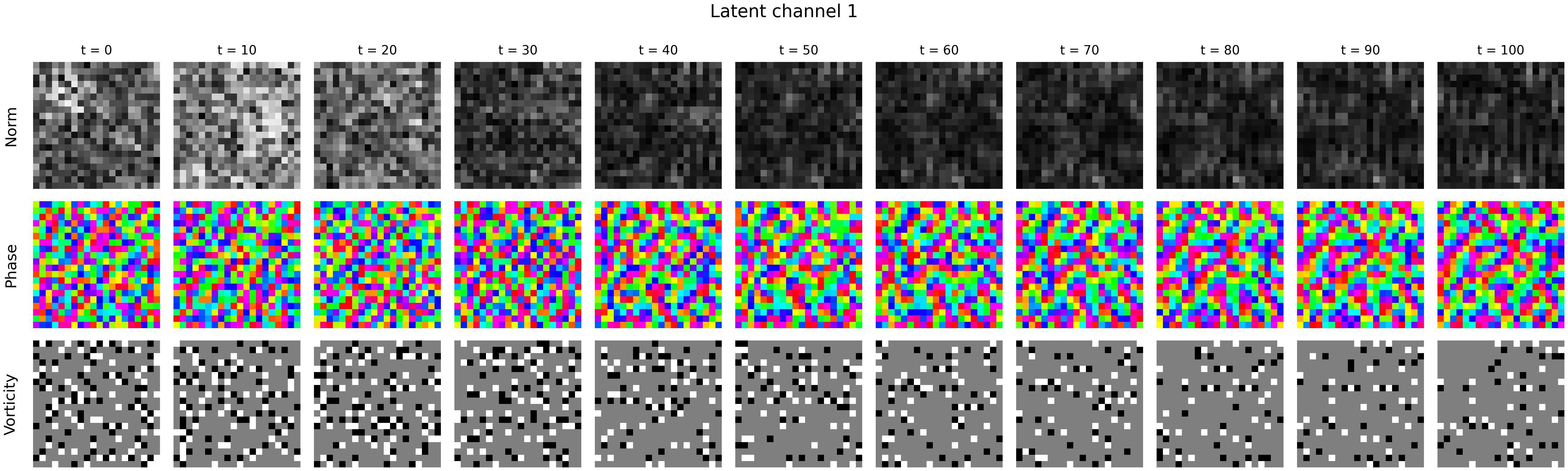}
    \caption{A second channel from the same 2D convolutional U(1) equivariant RNN. In this channel, we observe significantly higher vorticity that persists throughout the sequence length.}
    \label{fig:vortices_1}
\end{figure}

\newpage
\end{appendices}

\end{document}